\RequirePackage{fix-cm}
\documentclass[smallcondensed]{svjour3}     
\usepackage{booktabs}
\usepackage{amsmath}
\usepackage{amsfonts}
\usepackage[]{natbib} 
\usepackage{amssymb}
\usepackage{mathtools, stmaryrd}
\usepackage{xparse} \DeclarePairedDelimiterX{\Iintv}[1]{\llbracket}{\rrbracket}{\iintvargs{#1}}
\usepackage{xcolor}
\usepackage{colortbl}
\usepackage{url}
\usepackage{enumitem}

\usepackage{array}
\usepackage{multirow}

\usepackage[normalem]{ulem}
\usepackage{subfig}
\usepackage{graphicx}
\usepackage{rotating}

\newcommand{\Mat}{\textsf{\textup{M}}}  

\newcommand{\Y}{\mathcal{Y}}
\newcommand{\G}{\mathcal{G}}

\newcommand{\x}{\mathbf{x}}
\newcommand{\R}{\mathbb{R}}

\newcommand{\set}[1]{\left\{#1\right\}}

\newcommand{\vect}[1]{\overrightarrow{#1}}
\newcommand{\expectancy}[3]{ \operatornamewithlimits{\mathbb{E}^{\:#3}}_{\:#1} \left [ #2 \right ] } 
  
\newcolumntype{P}[1]{>{\centering\arraybackslash}p{#1}} 
 
\usepackage[]{algorithm}
\usepackage{algorithmic}

\usepackage{colortbl}
\usepackage{color}
\definecolor{my_blue}{rgb}{0.51,0.753,0.91}   
\definecolor{my_blue_lite}{rgb}{0.886,0.921,0.992}   
\definecolor{KO}{rgb}{0.992,0.941,0.886} 
\definecolor{OK}{rgb}{0.886,0.992,0.894} 

\usepackage{tabularx}

\smartqed  
\usepackage{graphicx}
\begin{document}

\title{Early Classification of Time Series 
}
\subtitle{Cost-based Optimization Criterion and Algorithms}

\titlerunning{Early Classification of Time Series}        

\author{	Youssef Achenchabe         \and
		Alexis Bondu		         \and
		Antoine Cornu\'ejols          \and
        		Asma Dachraoui	
}


\institute{Y. Achenchabe \at
              Orange Labs, 44 Avenue de la r\'epublique, Ch\^atillon, France \\
              UMR MIA-Paris, AgroParisTech, INRAe, 
              Universit\'e Paris-Saclay, 
              75005, Paris, France \\
              \email{youssef.achenchabe@universite-paris-saclay.fr}           
           \and
           A. Bondu \at
           Orange Labs, 44 Avenue de la r\'epublique, Ch\^atillon, France \\
           \email{alexis.bondu@orange.com}
           \and
           A. Cornu\'ejols \at
           UMR MIA-Paris, AgroParisTech, INRAe, 
           Universit\'e Paris-Saclay, 
           75005, Paris, France \\
           \email{antoine.cornuejols@agroparistech.fr}
           \and
           A. Dachraoui \at
           UMR MIA-Paris, AgroParisTech, INRAe, 
           Universit\'e Paris-Saclay, 
           75005, Paris, France \\
           \email{asma.dach@gmail.com}
}

\date{Received: 05-15-2020 / Accepted: date}

\maketitle

\vspace{-3mm}

\begin{abstract}
An increasing number of applications require to recognize the class of an incoming time series as quickly as possible  without unduly compromising the accuracy of the prediction. In this paper, we put forward a new optimization criterion which takes into account both the cost of misclassification and the cost of delaying the decision. Based on this optimization criterion, we derived a family of non-myopic algorithms which try to anticipate the expected future gain in information in balance with the cost of waiting. In one class of algorithms, unsupervised-based, the expectations use the clustering of time series, while in a second class, supervised-based, time series are grouped according to the confidence level of the classifier used to label them. 
Extensive experiments carried out on real datasets using a large range of delay cost functions show that the presented algorithms are able to solve the earliness vs. accuracy trade-off, with the supervised partition based approaches faring better than the unsupervised partition based ones. In addition, all these methods perform better in a wide variety of conditions than a state of the art method based on a myopic strategy which is recognized as being very competitive. Furthermore, our experiments show that the non-myopic feature of the proposed approaches explains in large part the obtained performances. 


\end{abstract}

\section{Introduction}

In emergency wards of hospitals \citep{mathukia2015modified}, in control rooms of national or international electrical power grids \citep{dachraoui2013early}, in government councils assessing emergency situations, in all kinds of contexts, it is essential to make timely decisions in absence of complete knowledge of the true outcome  (e.g. should the patient undergo a risky surgical operation?). 
The issue facing the decision makers is that, usually, the longer the decision is delayed, the clearer is the likely outcome (e.g. the critical or not critical state of the patient) but, also, the higher the cost that will be incurred if only because earlier decisions allow one to be better prepared. How to optimize online the tradeoff between the earliness and the accuracy of the decision is the object of the early classification of time series problem 
and this is what is addressed in this paper. 

The work presented takes the earliness versus accuracy trade-off at face value in the spirit of \citep{dachraoui2015early} and formalizes it in a generic way. In this paper, we extend this previous work as a generic framework for the early classification of time series. 

Furthermore, this previous work described a method based on the clustering algorithm K-means. However, the use of an unsupervised approach to capture the relevant groups of time series leaves important information aside. This is why, we propose here to resort to supervised techniques in order to get better prediction performances. 

Interestingly, we claim that the problem of deciding online whether a prediction, and the attendant actions, should be made, or if it should be delayed, can be cast in the LUPI (Learning Under Privileged Information) framework \citep{vapnik2009new}. During the learning phase, the learner has access to the full knowledge about the training time series in addition to their class, while at testing time, only the incoming, and incomplete, time series is known. And decisions as whether to wait for additional measurements or not have to be made from this incomplete knowledge. 
The resulting optimization criterion can be used in a \textit{non-myopic} procedure where estimates of future costs can be compared to the cost incurred if the decision was made at the current time. 

The second objective of this paper is to design efficient optimization algorithms which implement the presented framework.
We define a set of design choices that allow the categorization of possible \textit{non-myopic} approaches (see Table \ref{table:overview_methods}). This enables us to define three novel algorithms by varying these choices. They are then carefully evaluated and compared in experiments in order to identify the best approach. 
In addition, the three proposed \textit{non-myopic} approaches outperform the best known approach in the literature \citep{mori2017early} evaluated on the same collection of datasets proposed by the authors. 

The rest of the paper is organized as follows. The next section presents important works related to the early classification of time series problem.
Section \ref{sec_general_framework} reformulates in a generic way the cost-based formalism leading to an optimization criterion. 
Then, Section \ref{sec_approaches} presents the new methods that allow finer  estimations of this criterion. 
These methods vary depending on how they take advantage of the training set of complete time series to estimate future costs of decision. 
This gives rise to a set of questions as what are the characteristics that most drive the performance up. 
Section \ref{sec_expe} presents the experimental setup to answer these questions while the results with analyses are reported in Section \ref{sec_results}. Section \ref{sec_results_moris} compares the proposed approaches with the state of the art competing approach presented in  \citep{mori2017early}. 
Finally, Section \ref{sec_conclusion} concludes by underlying the main findings of this research and by discussing directions for future works. 

\section{{Related work}}
\label{related_work}

Formally, we suppose that measurements are made available over time in a time series which, at time $t$, is ${\mathbf x}_t \, = \, \langle {x_1}, \ldots, {x_t} \rangle$ where $x_t$ is the current measurement and the  ${x_i}_{(1 \leq i \leq t)}$ belong to some input domain (e.g. temperature and blood pressure of a patient). We suppose furthermore that each time series can be ascribed to some class $y \in {\cal Y}$ (e.g. patient who needs a surgical operation or not). The task is to make a prediction about the class of an incoming time series as early as possible because a cost is incurred at the time of the decision, where the cost function increases with time.  

\smallskip
If the measurements in a time series are supposed independently and identically distributed (i.i.d.) according to a distribution of unknown ``parameter'' $\theta$, then the relevant framework is the one of \textit{sequential decision making} and \textit{optimal statistical decisions} \citep{degroot2005optimal,berger1985statistical}. In this setting, 
the problem is to determine as soon as possible whether the measurements have been generated by a distribution of parameter $\theta_0$ (hypothesis $H_0$) or of parameter $\theta_1$ (hypothesis $H_1$) with $\theta_0 \neq \theta_1$. 
One technique especially has gained a wide exposition: Wald's Sequential Probability Ratio Test \citep{wald1948optimum}.  The log-likelihood ratio $R_t \; = \; \log \frac{P(\langle x_1^i, \ldots, x_t^i \rangle \; | \; y = -1)}{P(\langle x_1^i, \ldots, x_t^i \rangle \; | \; y = +1)}$ is computed and compared with 
two thresholds that are set according to the required error of the first kind $\alpha$ (\textit{false positive error}) and error of the second kind $\beta$ (\textit{false negative error}). Extensions to non-stationary distributions have been put forward (see \citep{liu2013performance,novikov2008optimal}). 

\smallskip
In the early classification of time series problem, however, the successive measurements are not supposed to be i.i.d. To compensate for this weaker assumption, it is assumed that a labeled training set exists made of time series of finite length $T$: ${\mathbf x}_T^i \, = \, \langle {x_1}^i, \ldots, {x_T}^i \rangle$ together with their corresponding labels, ${\cal S} = \{({\mathbf x}_T^i, y_i)\}_{1 \leq i \leq m}$. Each measurement $x_j^i$ can be multivariate.

In the test phase, the scenario goes as follows. At each time step $t < T$, a new measurement $x_t$ is collected and a decision has to be made as whether to make a prediction now or to defer the decision to some future time step. When $t = T$, a decision is forced.

As already mentioned, the problem of deciding online whether a prediction, and the attendant actions, should be made, or if it should be delayed, can be cast in the LUPI framework \citep{vapnik2009new}. 
In the following, we examine previous works on the early classification problem in this light. 

\smallskip
To the best of our knowledge, \citep{alonso2004boosting} is the earliest paper explicitly mentioning ``classification when only part of the series are presented to the classifier'', and the main thrust of it is to show how the boosting method can be employed to the classification of incomplete time series.

For many researchers, the question to solve is \textit{can we classify an incomplete times series while ensuring some minimum probability threshold that the same decision would be made on the complete input?} To answer this question several approaches have been put forward. 

One is to assume that the time series are generated i.i.d. according to some probability distribution, and to estimate the parameters of the class distributions from the training set. 
Once $p({\mathbf x}_T|{\mathbf x}_t)$ the conditional probability of the entire time series ${\mathbf x}_T$ given an incomplete realization ${\mathbf x}_t$ is estimated, it becomes possible to derive guarantees of the form: 

{
\begin{equation*}
p\bigl(h_T({\mathbf X}_T) = y|{\mathbf x}_t\bigr) \, = \, \int_{{\mathbf x}_T \text{ s.t. } h_T({\mathbf x}_T) = y} \, p({\mathbf x}_T|{\mathbf x}_t) \, d{\mathbf x}_T \, \geq \, \epsilon
\end{equation*}}where ${\mathbf X}_T$ is a random variable associated with the complete times series,  $\epsilon$ is a confidence threshold, and $h_T(\cdot)$ is a classifier learnt over the training set ${\cal S}$ of complete times series. At each time step $t$, $p(h_T({\mathbf X}_T) = y|{\mathbf x}_t)$ is evaluated and the prediction is triggered if this term becomes greater than some predefined threshold. 
\citep{anderson2012early,parrish2013classifying} present this method and propose ways to make the required estimations, in particular the mean and the covariance of the complete training data, when the time series are generated by Gaussian processes. It so far applies only with linear and quadratic classifiers.

\citep{xing2009early} do not make assumptions about the form of the underlying distributions on the time series. They propose to use a 1NN classifier that chooses the nearest training time series ${\mathbf x}_t^i \in {\cal S}$ to the incoming one ${\mathbf x}_t \, = \, \langle {x_1}, \ldots, {x_t} \rangle$ to make its prediction. To determine for which time step $t$ it is appropriate to make the prediction, the method is based on the idea of the \textit{minimum prediction length} (MPL) of a time series. For a time series ${\mathbf x}_t^i$, one finds the set of every training time series ${\mathbf x}_t^j$ that have ${\mathbf x}_t^i$ as their one nearest neighbor. The MPL of ${\mathbf x}_t^i$ is then defined as the smallest time index for which this set does not change when the rest of the time series ${\mathbf x}_t^i$ is revealed. 
In the test phase, at time step $t$, it is deemed that ${\mathbf x}_t$ can be safely labeled if its 1NN $= {\mathbf x}_t^i$ for which the MPL is $t$. The idea is that from this point on, the prediction about ${\mathbf x}_t$ should not change. The authors found experimentally that this procedure, called ECTS (\textit{Early Classification of Times Series}), leads to too conservative estimations of the earliest safe time step for prediction. They therefore proposed heuristic means to lower the estimated values. The stability criterion acts in a way as a proxy for a measure of confidence in the prediction. Similarly, \citep{mori2015early} proposes a method where the evolution of the accuracy of a set of probabilistic classifiers is monitored over time, which allows the identification of timestamps from whence it seems safe to make predictions.

{The authors of \citep{gupta2020early} are concerned with early classification of multivariate time series where the variables are not equally sampled in time. They focus on making a decision when all sensors have been processed and a high enough confidence level is attained. }

Another line of research is concerned with finding good descriptors of the time series, especially on their starting subsequences, so that early predictions can be reliable because they would be based on relevant similarities on the time series.  
For instance, in the works of \citep{xing2011extracting,ghalwash2014utilizing,he2015early}, the principle is to look for shapelets, subsequences of time series which can be used to distinguish time series of one class from another, so that it is possible to perform classification of time series as soon as possible. 

By contrast, there are methods for early classification of time series that do not take advantage of the complete knowledge available in the training set. For instance, in \citep{parrish2013classifying,hatami2013classifiers,ghalwash2012early}, a model $h_t(\cdot)$ is learnt for each early timestamp and various stopping rules are defined in order to decide whether, at time $t$, a prediction should be made or not. The price to pay for being outside the LUPI framework is that decisions are made in a myopic fashion which may prevent one from seeing that a better trade-off between earliness and accuracy is achievable in the future. 

This is also the case for the work presented in \citep{mori2019early}. 
In the paper, the authors recognize the conflict between earliness and accuracy, and instead of setting a tradeoff in a single objective optimization criterion \citep{mori2017early}
, they propose to keep it as a multi-objective criterion and to explore the Pareto front of the multiple dominating tradeoffs. Accordingly, they propose a family of triggering functions involving hyperparameters to be optimized for each tradeoff. 
This contrasts with approaches whereby the decision is made solely on the basis of a given confidence threshold which should be attained. However, the optimization criterion put forward is heuristic, supposes that the cost of delaying a decision is linear in time, and involves a complex setup. Most importantly, again, it is a myopic procedure which does not consider the foreseeable future. For all these apparent shortcomings, this method has been found to be quite effective, beating most competing methods in extensive experiments. This is why it is used as a reference method for comparison in this paper, as is done also in \citep{russwurm2019end} which compares several techniques for early classification of time series. 

The authors of \citep{schafer2020teaser} propose a system, called TEASER, that combines three components: (i) slave classifiers that estimate the class probabilities of the incoming series, (ii) a master classifier which assesses the confidence that one can have in the class that has the higher probability according to the slave classifier at the current time step, and finally (iii), the TEASER system which outputs a class if the master classifier has vetted this class for at least the last $v$ time steps. It is apparent that the cost of delaying decision is not explicitly taken into account. The authors propose to optimize the harmonic mean between accuracy and earliness which indirectly corresponds to a particular tradeoff and a particular cost of delaying the decision.  
Furthermore, the method is fairly empirical. 

In \citep{dachraoui2015early}, for the first time, the problem of early classification of time series is cast as the optimization of a loss function which combines the expected cost of misclassification at the time of decision plus the cost of having delayed the decision thus far. Besides the fact that this optimization criterion is well-founded, it permits also to apply the LUPI framework because the expected costs for an incoming subsequence ${\mathbf x}_t$ can be estimated for future time steps and thus a non-myopic decision procedure can be used. These expectations can indeed be learned from the training set of $m$ complete time series ${\cal S} = \{({\mathbf x}_T^i, y_i)\}_{1 \leq i \leq m}$. 

\section{Early classification as a cost optimization problem}
\label{sec_general_framework}

Classifying time series with as little measurements as possible implies optimizing a trade-off. The less data is available for classification, the lower is the attainable accuracy in general. But waiting for more data implies incurring higher delay costs. There is therefore an optimization problem to be solved involving both the classification accuracy, which translates into a misclassification cost, and the cost associated with gathering measurements. In the framework of online classification, this optimization problem becomes an online sequential decision making problem, where at each new time step, and a corresponding new piece of data, the system must decide whether to output a label for the incoming time series or to wait for more measurements. If the decision is made solely on the basis of the currently available information, the approach is myopic. On the other hand, if the decision involves some sort of prediction about the expected value of the optimization criterion in the future, the approach is non-myopic. This is what is allowed by the LUPI framework. Such a perspective was presented in \citep{dachraoui2015early}. We generalize it in this section and lay the ground for three novel cost-based optimization criteria that are the object of this paper.  

\smallskip

This approach assumes that the user provides two cost functions: 
\begin{itemize}
	\item $\mathrm{C}_m(\hat{y}|y) : {\cal Y} \times {\cal Y} \rightarrow \R$ is the \textit{misclassification cost function} that defines the cost of predicting $\hat{y}$ when the true class is $y$.
	\item $C_d(t) : \R \rightarrow \R$ is the \textit{delay cost function} which is non decreasing over time.
\end{itemize}

Both of these costs are expressed in the same unit (e.g. in dollars) and convey the characteristics of the application domain as known by experts. 

The expected cost of a decision at time step $t$, when ${\mathbf x}_t$ is the incoming time series, can be expressed as:

\begin{align}
\begin{split}
\normalsize
\label{eq:cost1}
f(\mathbf{x}_t) \; &= \; \expectancy{}{\mathrm{C}_m|x_t}{t} \; +  \mathrm{C}_d(t) \;  = \; \sum_{(y,\hat{y}) \in {\cal Y}^2}  P_{t}(\hat{y},y|x_t) \mathrm{C}_m(\hat{y}|y)  + \; \mathrm{C}_d(t) \\
 &= \; \sum_{y \in {\cal Y}} P_{t}(y|\mathbf{x}_t) \, \sum_{\hat{y} \in {\cal Y}} P_t(\hat{y}|y, \mathbf{x}_t) \, \mathrm{C}_m(\hat{y}|y) \; + \; \mathrm{C}_d(t)
\end{split}
\end{align}

The expectation comes both from the misclassification probability {$P_t(\hat{y}|y, \mathbf{x}_t)$ which can be estimated by the confusion matrix of the classifier $h_t(\cdot)$ applied at time $t$, and the posterior probability of each class given the input incomplete time series estimate $P_t(y|\mathbf{x}_t)$. 

If the input time series was fully observed, this cost could be computed for all time steps $t \in \{1, \ldots, T\}$, and the optimal time $t^*$ for triggering the classifier's prediction would be: 
{
\begin{equation}
\label{eq:time1}
t^* \; = \; \operatornamewithlimits{ArgMin}_{t \in \{1, \ldots, T\}} f(\mathbf{x}_t)
\end{equation}}

But of course, this would defeat the whole purpose of early classification, as one would have to observe the entire time series before knowing what would have been the optimal decision time!
Then, instead of waiting until the entire time series is known, at each time $t$, one could ``look into the future'' and guess what will be the best decision time.  
And if the estimated best decision time matches the current time step $t$, then the decision must be made.  
For the incoming time series $\x_t$, the expected cost at $\tau$  time steps in the future is:
\begin{equation} 
\label{eq:cost2}
f_{\tau}(\x_t)  = 
   \sum_{y \in {\Y}} P_{t+\tau}(y|\x_t) 
 \sum_{\hat{y} \in {\Y}} P_{t+\tau}(\hat{y}|y,\x_{t+\tau}) \, \mathrm{C}_m(\hat{y}|y) 
   + \mathrm{C}_d(t+\tau)
\end{equation}}
where $\x_{t+\tau}$ is the foreseen continuation of $\x_t$. Accordingly, the best expected decision time in the future becomes:
{
\begin{equation}
\label{eq:time2}
\tau^* \; = \; \operatornamewithlimits{ArgMin}_{\tau \in \{0, \ldots, T-t\}} f_\tau(\mathbf{x}_t)
\end{equation}
and  if $\tau^* = 0$ the decision is instantly requested, and $\widehat{t^*}=t$ denotes the trigger time.
The problem now is \textit{how to predict $\x_{t+\tau}$ from the knowledge of $\x_t$}. Can the LUPI framework help? Yes it can. Figure \ref{fig-foreseable-futures-1} provides an overview of the principle in the case of a univariate time series. The ``envelope'' of its foreseeable futures can be learned using the training dataset of complete time series ${\cal S} = \{({\mathbf x}_T^i, y_i)\}_{1 \leq i \leq m}$. 

Importantly, the solution chosen to guess the ``envelope'' of the $\x_{t+\tau}$ will also provide a way to estimate the terms $P_{t+\tau}(\hat{y}|y,\x_{t+\tau})$ because a confusion matrix can be learned on this envelope. 

However, estimating the likely outcomes of the incoming time series using a probabilistic forecasting model involves making assumptions (e.g. assuming that the residuals distribution is Gaussian). 
Another way to facilitate the use of this cost-based formalism is taken in \citep{dachraoui2015early} which consists in learning typical groups of time series from the training set, and then in predicting the likely continuations of $\x_t$ with regard to these groups (see Figure \ref{fig-foreseable-futures-2}). 

\begin{figure}[htbp!]
\centering
\subfloat[]{\label{fig-foreseable-futures-1} \includegraphics[width=0.49\linewidth]{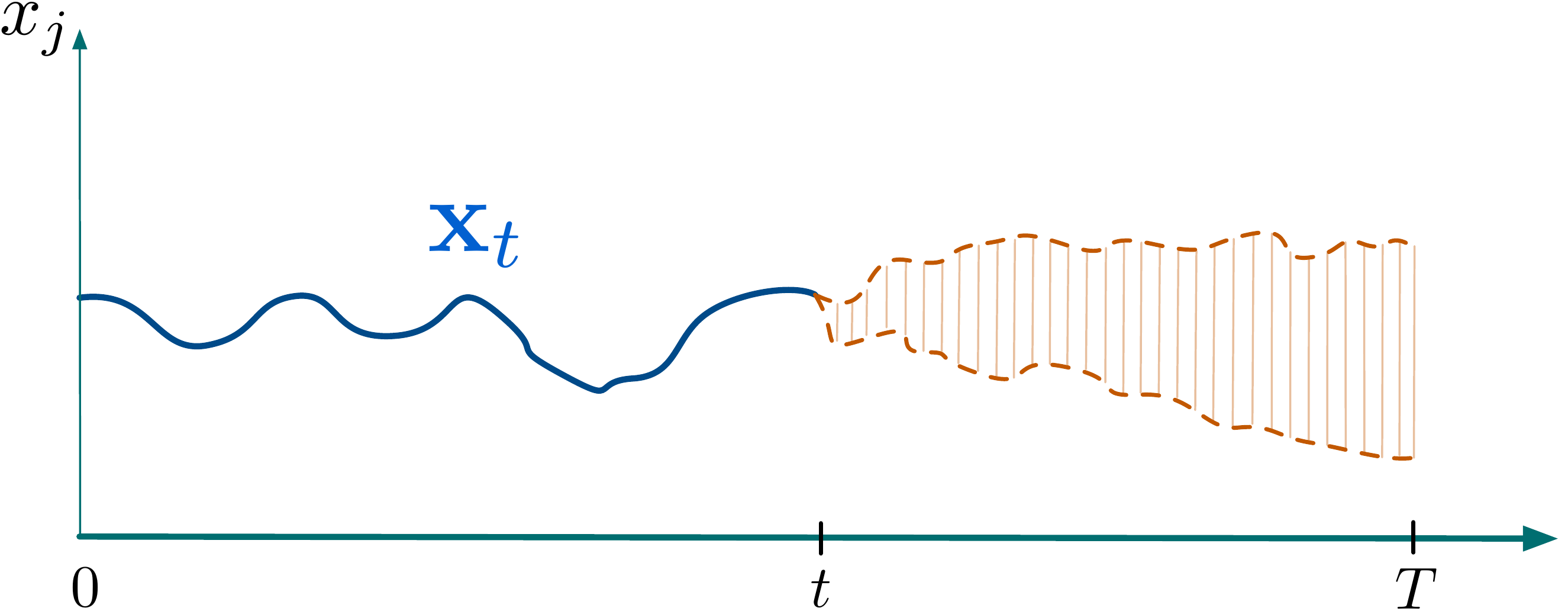}}
\subfloat[]{\label{fig-foreseable-futures-2} \includegraphics[width=0.49\linewidth]{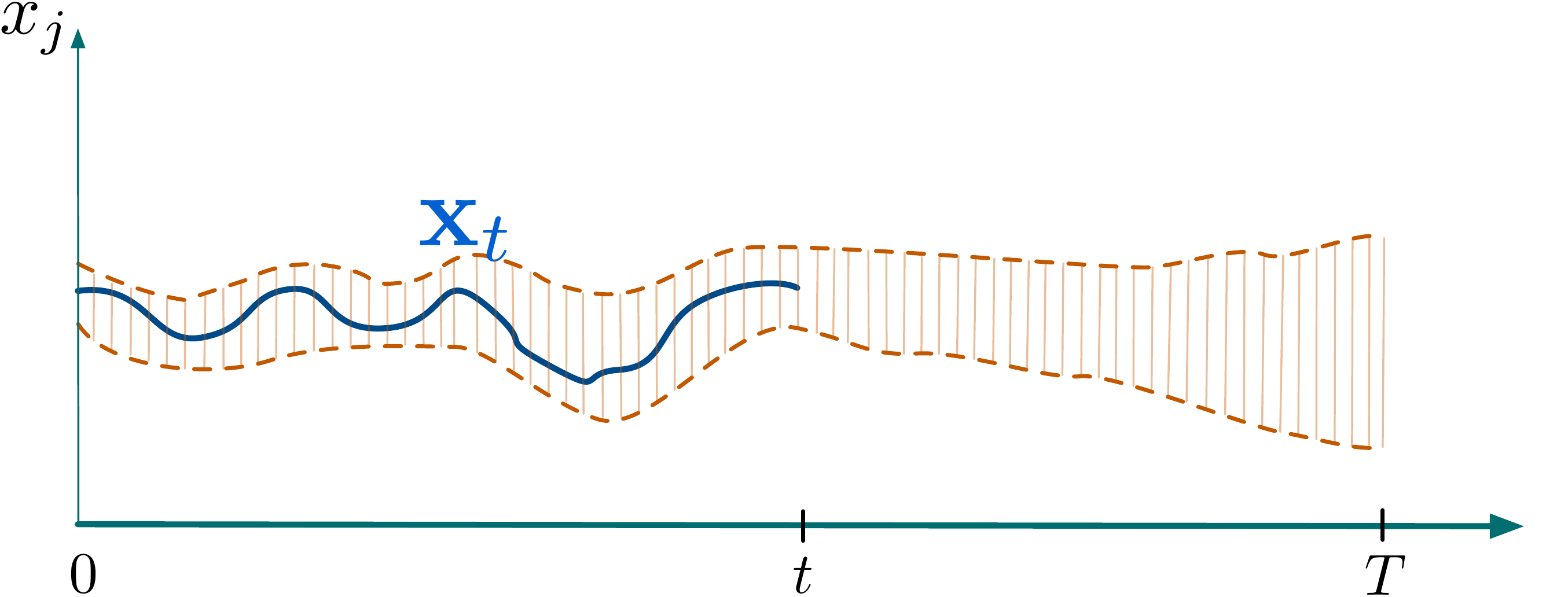}} 
\caption{(a) Given an incomplete time series $\x_t$, the objective is to try to guess the ``envelope'' of its foreseeable futures. Various methods can be used to do so. (b) The incoming time series $\x_t$ is viewed as a member of or close to some group(s) of times series, and this is used to guess the ``envelope'' of its foreseeable futures.}
\label{fig-foreseable-futures}
\end{figure}

Let us note $\mathfrak{g}_k$ the $k$-th typical groups of time series, Equation (\ref{eq:cost1}) then can be re-expressed as:
{
\begin{equation}
\label{eq:cost3}
f(\mathbf{x}_t)=
\sum_{\mathfrak{g}_k \in \G} P_{t}(\mathfrak{g}_k|\x_t)  \sum_{y \in {\Y}} P_{t}(y|\mathfrak{g}_k)  
 \sum_{\hat{y} \in {\Y}} P_{t}(\hat{y}|y,\mathfrak{g}_k) \mathrm{C}_m(\hat{y}|y) 
+ \mathrm{\mathrm{C}}_d(t)
\end{equation}}

And similarly, for Equation (\ref{eq:cost2}):

{
\begin{equation}
\label{eq:cost4}
f_{\tau}(\x_t)  = 
  \sum_{\mathfrak{g}_k \in \G} P_{t}(\mathfrak{g}_k|\x_t) \sum_{y \in {\Y}} P_{t}(y|\mathfrak{g}_k) 
 \sum_{\hat{y} \in {\Y}} P_{t+\tau}(\hat{y}|y,\mathfrak{g}_k) \, \mathrm{C}_m(\hat{y}|y) 
   + \mathrm{C}_d(t+\tau)
\end{equation}}

Equation (\ref{eq:cost4}) can be easily interpreted by splitting it into two parts.  
The first term $P_t(\mathfrak{g}_k|\x_t)$ estimates the posterior probabilities of each group given $\x_t$.
{This term is estimated at time $t$ and assumed to be constant over the time interval $[t,t+\tau]$.}
The next term expresses the expectations of the cost of misclassification over future possible continuations of $\x_t$. 
Namely, the second term $P_t(y|\mathfrak{g}_k)$ corresponds to the prior probabilities of class values within each group {estimated at time $t$}. 
And the third term $P_{t+\tau}(\hat{y}|y,\mathfrak{g}_k)$ estimates the probabilities of misclassification within each group, at time step $t + \tau$. 
The terms $\mathrm{C}_m(\hat{y}|y)$ and $\mathrm{C}_d(t+\tau)$ are the cost functions expressing properties of the domain of application.   

In this general framework, several choices can be made to implement this optimization criteria. Foremost is the determination of relevant groups $\mathfrak{g}_k$ of time series from the complete training set ${\cal S}$. 
In what follows, we propose four different alternatives to anticipate the expected future misclassification costs.  

\section{Anticipating the future: a key to the optimization criterion}
\label{sec_approaches}

This section presents three novel non-myopic approaches to solve cost-based optimization problem. These are three ways of Learning Using Privileged Information (LUPI) and therefore being able to foresee likeliest future values for the optimization criterion. The characteristics of these approaches are summarized in Table \ref{table:overview_methods}.

\vspace{-5mm}
 
\begin{center}
\begin{table}[!h]
\scriptsize
\begin{tabularx}{\textwidth}{| X | X | X | X |}
 \hline 
\cellcolor{my_blue} Approaches 	& \cellcolor{my_blue} Partition type 	&  \cellcolor{my_blue} Partitions number		&	\cellcolor{my_blue} Anticipation type	 \\
\hline
\hline
 \cellcolor{my_blue_lite} \textsc{Economy-K} {\tiny \citep{dachraoui2015early}} 		&	\cellcolor{KO} {\bf Unsupervised} \newline \textit{(K-means)} 	&  \cellcolor{KO}{\bf 1} partition of full-length time series.  	&	\cellcolor{KO}{\bf Weak model}: the continuations of time series are used in the groups.	 \\
\hline
\cellcolor{my_blue_lite} \textsc{Economy-multi-K} 		& \cellcolor{KO} {\bf Unsupervised} \newline \textit{(K-means)} 	&  \cellcolor{OK}{\bf T} partitions corresponding to the different time steps.  	&	\cellcolor{KO}{\bf Weak model}: the continuations of time series are used in the groups.	 \\
\hline

\cellcolor{my_blue_lite} \textsc{Economy-$\gamma$-lite}  \textit{\tiny (binary classification)} 		&	\cellcolor{OK}{\bf Supervised}: quantiles of the classifiers' confidence levels are used  	&  \cellcolor{OK}{\bf T} partitions corresponding to the different time steps.  	&	\cellcolor{KO}{\bf Weak model}: the continuations of time series are used in the groups.	 \\
\hline

\cellcolor{my_blue_lite} \textsc{Economy-$\gamma$}   ~~~~~~~~~~~~~~~~\textit{\tiny (binary classification)}		&	\cellcolor{OK}{\bf Supervised}: quantiles of the classifiers' confidence levels are used 	&  \cellcolor{OK}{\bf T} partitions corresponding to the different time steps.  	&	\cellcolor{OK}{\bf Markov Chain} technique is used to anticipate missing measurements. 	 \\
\hline

\end{tabularx}
\caption{Overview of the design choices of the different approaches: each approach differing from the previous one by only one design choice. \label{tab:approaches}}
\label{table:overview_methods}
\end{table}
\end{center}
 
\vspace{-8mm} 

The three proposed approaches seek to better extract information about the foreseeable future of incoming time series so as to offer a better basis for deciding whether to label the time series at the current time step or to wait for more measurements. As can be seen from Table \ref{table:overview_methods}, each proposed approach can be viewed as an incremental modification of a previous method, from \textsc{Economy-K} presented in \citep{dachraoui2015early} to \textsc{Economy-$\gamma$}.

Where \textsc{Economy-K} is computing partitions of the time series based on the complete training time series, through a clustering process, \textsc{Economy-multi-K} partitions incomplete times series for each time step in order to increase adaptiveness to the incoming and incomplete time series (see Section \ref{Economy-multi-K}). Both methods do not use the labels of the time series in order to compute the partitions. 

By contrast, both \textsc{Economy-$\gamma$-lite} and \textsc{Economy-$\gamma$} use the labels in order to predict the likely future of the incoming time series. \textsc{Economy-$\gamma$-lite} uses the level of confidence of the classifier at time $t$ on the incoming ${\mathbf x}_t$ in order to define groups of time series (see Section \ref{Economy-gamma-lite}), whereas \textsc{Economy-$\gamma$} uses Markov chains in order to anticipate the likely future measurements on the time series (see Section \ref{Economy-gamma}). The current implementations of \textsc{Economy-$\gamma$-lite} and of \textsc{Economy-$\gamma$}
 only accommodate binary classification tasks, but extensions to multi-class problems are envisioned for future work.

The number of groups $K$ is a hyper-parameter shared by all of these approaches. In practice, it can be tuned using cross validation as detailed in Section \ref{experimental_protocol}.  
An open-source code is available for full reproducibility of the experiments presented in this paper: {\url{https://github.com/YoussefAch/Economy}}. 
The following sub-sections provide the main operating principles and key ideas for each \textsc{Economy} approach.

\vspace{-4mm}

\subsection{\textsc{Economy-K}}
\label{Economy-K}

\vspace{-2mm}

\textsc{Economy-K} has been introduced in \citep{dachraoui2015early}. The idea is to first identify groups $\mathfrak{g}_k$ of times series using a clustering algorithm, here K-means with Euclidian distance.  
Then, given an incoming time series $\x_t$, the memberships $P(\mathfrak{g}_k|\x_t)$ are estimated using a logistic function of a distance between $\x_t$ and the centers of the clusters $\mathfrak{g}_k$. 
In order to estimate the terms $P_{t}(\hat{y}|y,\mathfrak{g}_k)$  of the confusion matrix for each time step $t = 1, \ldots, T$, 
a collection of classifiers $\{h_t\}_{t \in \{1, \ldots, T\}}$ is learned using training sets $\{{\cal S}^t\}_{t \in \{1, \ldots, T\}}$ of time series truncated to their first $t$ measurements. 

In the end, implementing the \textsc{Economy-K} approach relies on the following choices:  (i) a distance function for K-means, (ii) a distance between an incomplete time series and clusters, (iii) a membership function estimating $P(\mathfrak{g}_k|\x_t)$.

As explained in Section \ref{sec_general_framework}, Equation (\ref{eq:cost4}) is used to estimate the cost of deciding for future time steps $t + \tau$  ($0 \leq \tau \leq T - t$), and if $\tau^*$ given by Equation (\ref{eq:time2}) is equal to zero or $t = T$, then a decision is triggered, otherwise a new measurement $x_{t+1}$ is made, and the decision mechanism is called again. 

\vspace{-3mm}

\begin{algorithm}[h]
\begin{algorithmic}[1]
{\small
    \REQUIRE K: number of groups
    \STATE build the groups $\mathfrak{g}_k$ by using K-means on a set of full-length time series ${\cal S}$    
 	\FORALL{$t \in \set{1, \dots, T}$} 
		\STATE learn a classifier $h_t(\cdot)$ from a set of truncated time series ${\cal S}^t$
		\FORALL{$g_k \in \G$}
		\STATE estimate the confusion matrix of $h_t(\cdot)$ in the group $g_k$
		\ENDFOR	
	\ENDFOR	
}
\end{algorithmic}
\caption{\small Learn \textsc{Economy-K}}
\label{algo:learn_ECO_K}
\end{algorithm}

\vspace{-3mm}

Algorithm \ref{algo:learn_ECO_K} provides the pseudo-code which summarizes the learning stage of the  \textsc{Economy-K} approach. The next sections describe the algorithms \ref{algo:learn_ECO_multi_K}, \ref{algo:learn_gamma_lite} and \ref{algo:learn_gamma} emphasizing for each of them the single difference with the previous  algorithm presented (see the italicized bold line in each algorithm).
  
\vspace{-3mm}

\begin{algorithm}[h]
\begin{algorithmic}[1]
{\small
    \REQUIRE K: number of groups
 	\FORALL{$t \in \set{1, \dots, T}$} 
		\STATE {\bf \textit{build the groups $\mathfrak{g}_k^t$ by using K-means on a set of truncated time series ${\cal S}^t$}} 
		\STATE learn a classifier $h_t(\cdot)$ from a set of truncated time series ${\cal S}^t$
		\FORALL{$g_k \in \G$}
		\STATE estimate the confusion matrix of $h_t(\cdot)$ in the group $g_k$
		\ENDFOR	
	\ENDFOR	
}
\end{algorithmic}
\caption{\small Learn \textsc{Economy-multi-K}}
\label{algo:learn_ECO_multi_K}
\end{algorithm}

\vspace{-7mm}

\subsection{\textsc{Economy-multi-K}}
\label{Economy-multi-K}

\vspace{-2mm}

Instead of grouping time series using their full-length descriptions, an alternative consists in computing the clusters $\mathfrak{g}_k^t$ for each time step $t$ using training sets ${\{\cal S}^t\}_{t \in \{1, \ldots, T\}}$ of truncated time series from the training set ${\cal S}$ {(see line 2 of Algorithm \ref{algo:learn_ECO_multi_K})}. 
Indeed, clustering time series on the fly, at each time step, may allow for a increased adaptiveness to the specifics of the the beginning of the time series. The term $P(\mathfrak{g}_k|\x_t)$ in Equation \ref{eq:cost4} then becomes $P(\mathfrak{g}_k^t|\x_t)$. The cost of potential future decisions is now estimated based on the terms $P_{t+\tau}(\hat{y}|y,\mathfrak{g}_k^t)$. 

\vspace{-4mm}

\subsection{\textsc{Economy-$\gamma$-lite}}
\label{Economy-gamma-lite}

\vspace{-2mm}

In the previous approaches, the confusion matrix with the term $P_{t+\tau}(\hat{y}|y,\mathfrak{g}_k)$ in Equation (\ref{eq:cost4}), is computed using time series in $\mathfrak{g}_k$ and potentially aggregates all confidence levels of $h_{t + \tau}$, corresponding to all possible values of  the conditional probability $p(y=1|{\mathbf x}_{t+ \tau})$. If this confusion matrix was instead computed over time series that share approximately the same confidence level in their classification, the estimation of future decision costs could be much more precise. This is the motivation behind the algorithms \textsc{Economy-$\gamma$} and \textsc{Economy-$\gamma$-lite}. 

In these methods, the groups $\mathfrak{g}_k^t$ are obtained by stratifying the time series by confidence levels\footnote{This restricts these methods to binary classification problems.} of $h_t$ {(see line 3 of Algorithm \ref{algo:learn_gamma_lite})}.
At each time step $t$, the confidence level $p(h_t(\x_t) = 1)$ of the classifier can take a value in $[0,1]$. Examining the confidence 
levels for all time series in the validation set ${\cal S'}^t$ truncated to the first $t$ observations, we can discretize 
the interval $[0, 1]$ into $K$ equal frequency intervals, denoted} $\{I_t^1,\ldots, I_t^K\}$. For instance, if $K=5$, and $|{\cal S'}^t| = 1000$, the intervals  $I_t^1 = [0, 0.30[$, $I_t^2 = [0.30, 0.45[$, $I_t^3 = [0.45, 0.58[$, $I_t^4 = [0.58, 0.83[$, $I_t^5 = [0.83, 1]$ could each correspond to 200 training time series. The discretization of confidence levels into equal frequency intervals corrects any bias in the calibration of $h_t$, in a similar way to isotonic calibration \citep{flach2016classifier}.

Given an incoming time series ${\mathbf x}_t$, the classifier $h_t$ is used to get an estimate of $p(y=1|{\mathbf x}_t)$ and determine the group $\mathfrak{g}_k^t$ to which ${\mathbf x}_t$ belongs.
The algorithm is the same as \textsc{Economy-multi-K}, only with the groups $\mathfrak{g}_k^t$ obtained in a supervised way by leveraging the information about the membership to the classes.

One can notice that,  in addition to the expected gain in performance due to a more informed grouping of time series than in the clustering-based approaches, this method as well as \textsc{Economy-$\gamma$}, does not require (i) the choice of a distance function for K-means, nor (ii) the determination of another distance between an incomplete time series $\x_t$ and a cluster of full-length time series, and finally (iii) neither the choice of a membership function in order to estimate $P(\mathfrak{g}_k|\x_t)$. The approach is therefore much simpler to implement.

\vspace{-5mm}

\begin{figure}[htbp!]
\centering
\includegraphics[width=0.9\linewidth]{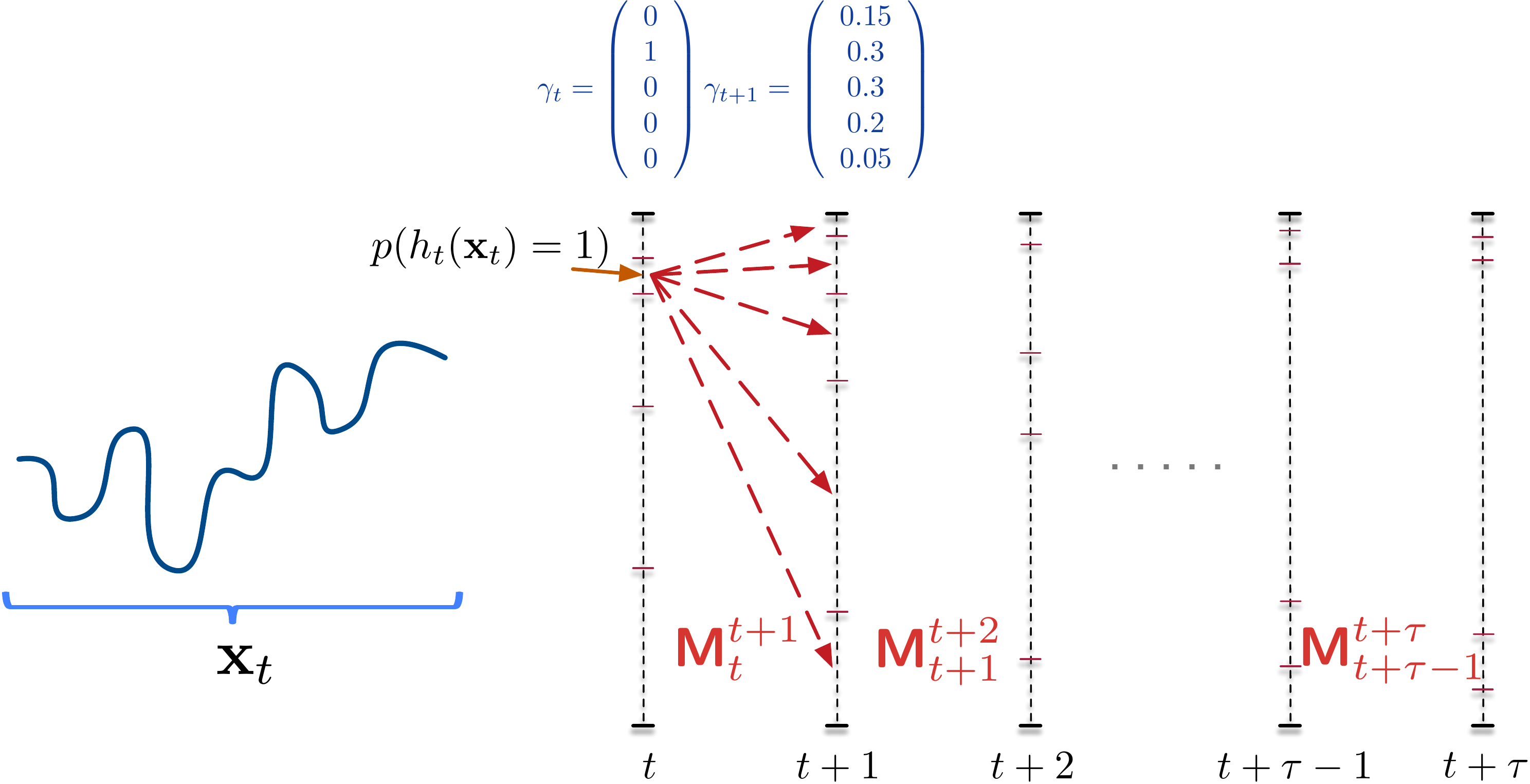}
\caption{\textsc{Economy-$\gamma$}, computing the probability distribution $p(\gamma_{t+\tau}|\gamma_t)$. Here $h_t(\x_t)$ falls in the second confidence level interval. Given a supposed learned transition matrix $\Mat_{t}^{t+1}$, the next vector of confidence levels will be 
${(0.15, 0.3, 0.3, 0.2, 0.05)^{\top}}$. 
}
\label{fig-confidence-intervals-new}
\end{figure}

\begin{algorithm}[h]
\begin{algorithmic}[1]
{\small
    \REQUIRE K: number of groups
 	\FORALL{$t \in \set{1, \dots, T}$} 
		\STATE learn a classifier $h_t(\cdot)$ from a set of truncated time series ${\cal S}^t$
		\STATE {\bf \textit{build the groups $\mathfrak{g}_k^t$ by discretizing the confidence level of $h_t(\cdot)$ into $K$ intervals}}
		\FORALL{$g_k \in \G$}
		\STATE estimate the confusion matrix of $h_t(\cdot)$ in the group $g_k$
		\ENDFOR	
	\ENDFOR	
}
\end{algorithmic}
\caption{\small Learn \textsc{Economy-$\gamma$-lite}}
\label{algo:learn_gamma_lite}
\end{algorithm}

\subsection{\textsc{Economy-$\gamma$}}
\label{Economy-gamma}

\vspace{-2mm}

\textsc{Economy-$\gamma$} uses the \textsc{Economy-$\gamma$-lite} principle to assign an incoming time series $\x_t$ to a given group $\mathfrak{g}_k^t$, but it tries to get better estimates of the future terms $P_{t+\tau}(\hat{y}|y,\mathfrak{g}_k^t)$ of the confusion matrices by replacing $\mathfrak{g}_k^t$ by a projection $\mathfrak{g}_k^{t+\tau}$ into the future as a probability distribution over the confidence intervals of $h_{t + \tau}$. 

Let us call $\vect{\gamma_{t}} = (\gamma_t^1, \ldots, \gamma_t^K)^{\top}$ the real-value vector of $K$ components $\gamma_t^i$, where each of the components is the probability that $p(h_{t}(\x_t) \in I_t^i)$. For instance, in Figure \ref{fig-confidence-intervals-new}, $\vect{\gamma_{t}} = (0, 1, 0, 0, 0)^{\top}$, all components are zero except $\gamma_t^2 = 1$.  

We would like to compute the vectors $\vect{\gamma_{t + \tau}}$ ($0 < \tau \leq T-t)$ consisting of the components: 

\vspace{-4mm}

{
\begin{equation}
\gamma^j_{t + \tau} \;  = \; p(h_{t+\tau}(\x_{t + \tau}) \in I_{t+\tau}^j) 
\end{equation}}

In \textsc{Economy-$\gamma$}, we propose to estimate $\gamma^j_{t + \tau}$ by using the $K \times K$ transitions matrices $\{\Mat_{t'}^{t'+1}\}_{t' \in \{1, \ldots, T-1\}}$ from $\vect{\gamma_{t'}}$ to $\vect{\gamma_{t'+1}}$ {(see line 4 of Algorithm \ref{algo:learn_gamma})} where each component of the matrix is estimated by:

\vspace{-2mm}

{
\begin{equation}
m_{i,j} \;  = \; p ( \, h_{t'+1}(\x_{t'+1}) \in I_{t'+1}^j \, \, | \, \, h_{t'}(\x_{t'}) \in I_{t'}^i \,) 
\end{equation}}
given a validation set of time series.
At time step $t$, and from $\vect{\gamma_t}$ it then becomes possible to compute $\vect{\gamma_{t + \tau}}$ by:

\vspace{-4mm}

{
\begin{equation}
\vect{\gamma_{t + \tau}} \; = \; \vect{\gamma_t}^{\top} \prod_{s=0}^{\tau-1} \Mat_{t+s}^{t+s+1}  
\label{eq:gamma-future-state}
\end{equation}}
Like in Equation (\ref{eq:cost4}), the future expected costs of decision are estimated through:
{
\begin{equation}
f_{\tau}(\x_t) \;  = \; \underbrace{\sum_{j = 1}^K \gamma_{t + \tau}^j}_{(1)} \, 
\sum_{y \in {\cal Y}} P(y|I_{t + \tau}^j) \, 
\underbrace{ \vphantom{\sum_{j = 1}^K }
\sum_{\hat{y} \in {\Y}} P_{t + \tau}(\hat{y} | y, I_{t + \tau}^j) }_{\text{(2)}} 
\, \mathrm{C}_m(\hat{y}|y)
+ \mathrm{\mathrm{C}}_d(t+\tau) 
\label{eq:cost-new}
\end{equation}} 

\noindent
(1): for all confidence intervals $I_{t+\tau}^j$ of $h_{t + \tau}$   \\
(2): probability of misclassification when $h_{t + \tau}(\x_{t+\tau}) \in I_{t+\tau}^j$ 

\smallskip
Again, a decision is triggered at time $\widehat{t^*}=t$, if $\tau^* \; = \; \operatornamewithlimits{ArgMin}_{\tau \in \{0, \ldots, T-t\}} f_\tau(\mathbf{x}_t)$ is found to be 0.

\begin{algorithm}[h]
\begin{algorithmic}[1]
{\small
    \REQUIRE K: number of groups
 	\FORALL{$t \in \set{1, \dots, T}$} 
		\STATE learn a classifier $h_t(\cdot)$ from a set of truncated time series ${\cal S}^t$
		\STATE build the groups $\mathfrak{g}_k^t$ by discretizing the confidence level of $h_t(\cdot)$ into $K$ intervals
		\STATE {\bf \textit{estimate the transition matrix  $\Mat_{t}^{t+1}$}}
	\ENDFOR	
}
\end{algorithmic}
\caption{\small Learn \textsc{Economy-$\gamma$}}
\label{algo:learn_gamma}
\end{algorithm}

\vspace{-10mm}

\subsection{{Complexity analysis}}

We provide here an analysis of the computational complexities of the proposed algorithms, first in relation to the learning stage, and then with regard to the inference phase. 

These complexities are expressed in a generic way, i.e. regardless of the classifier (of complexity denoted \textit{Learn}) and clustering (\textit{Clustering}) algorithms used. The single difference between two successive algorithms is underlined using bold letters in the following expressions that relate to the \textbf{learning phase}.

{\small
\begin{itemize}[itemindent=7em]
    \item [\textsc{Eco-K}:] $\mathcal{O}(T.Learn + Clustering +  T.|\mathcal{S}|.Predict + |Y|.K.|\mathcal{S}|)$ 
\smallskip
    \item [\textsc{Eco-multi-K}:] $\mathcal{O}(T.Learn +  {\bf T}.Clustering  +  T.|\mathcal{S}|.Predict+|Y|.K.|\mathcal{S}|)$ 
\smallskip
    \item [\textsc{Eco-$\gamma$-lite}:] $\mathcal{O}(T.Learn + T.{\bf |\mathcal{\bf S}|.log(|\mathcal{\bf S}|})) +  T.|\mathcal{S}|.Predict + |Y|.K.|\mathcal{S}|)$ 
\smallskip
    \item [\textsc{Eco-$\gamma$}:] $\mathcal{O}(T.Learn + T.|\mathcal{S}|.log(|\mathcal{S}|)) +  T.|\mathcal{S}|.Predict + |Y|.K.|\mathcal{S}| + |\mathcal{\bf S}|^{\bf2} . {\bf K^2})$   
\end{itemize}
}

\textsc{Economy-K} learns a collection of $T$ classifiers and build a partition of full-length time series using a $Clustering$ algorithm with a  $\mathcal{O}(T.Learn + Clustering)$ complexity. Then, confusion matrices are computed with a $\mathcal{O}(T.|\mathcal{S}|.Predict)$ complexity and the prior probability for each class in each group is computed with a $\mathcal{O}(|Y|.K.|\mathcal{S}|)$ complexity. 

\textsc{Economy-multi-K} has almost the same complexity and only differs by computing ${\bf T}$ different partitions. 

\textsc{Economy-$\gamma$-lite} discretizes the outputs of the classifiers by sorting time series according to their confidence level with a $\mathcal{O}(T.{\bf |\mathcal{\bf S}|.log(|\mathcal{\bf S}|}))$ complexity. 

Finally, \textsc{Economy-$\gamma$} adds the estimation of the transition matrices computed with a $\mathcal{O}(|\mathcal{\bf S}|^{\bf 2} . {\bf K^2})$ complexity.
  
\smallskip
During the \textbf{inference phase}, a new measurement is received at each time step and the future costs must be estimated. All the  approaches carry out this estimate in a $\mathcal{O}(T.|Y|^2.K)$ complexity, except \textsc{Economy-$\gamma$} that uses a matrix-vector product for transition matrix to estimate future costs with a $\mathcal{O}(T.|Y|^2.K + K^2)$ time complexity. 

\vspace{-4mm}
\section{Description of the experiments}
\label{sec_expe}
\vspace{-2mm}

\subsection{Goal of the experiments}
\label{goal_expe}
\vspace{-2mm}

The approaches presented all rely on the estimation of the best decision time based according to a cost-based criterion which expresses the expected misclassification cost for future time steps plus a delay cost. This is the basis of these non-myopic strategies. 

A \textit{first set of questions} regards the importance and \textit{impact of the various design choices} that distinguish the four \textsc{Economy} algorithms (see first part of the experiments in Section \ref{expe_ECO_only}):

\begin{enumerate}
    \item Time series are partitioned in an unsupervised way for \textsc{Economy-K} and \textsc{Econo\-my-Multi-K} and in a supervised mode for \textsc{Economy-$\gamma$-lite} and \textsc{Economy-$\gamma$}. Is one approach better than the other? 
    \item On a finer grain, is it better to cluster series using their full-length descriptions as in {\textsc{Economy}-K} or on their truncated description at each time step $t$  as in {\textsc{Economy}-multi-K}?
    \item Is it useful to try to have a more precise anticipation of the future of the incoming time series as is done in \textsc{Economy-$\gamma$} compared with the simpler albeit coarser approach of \textsc{Economy-$\gamma$-lite}? 
    \item How distant the cost incurred is from the ideal optimal cost that one would have paid if one had known the whole series and therefore the best decision time? This is akin to a \textit{regret} for not having a perfect a posteriori knowledge.
\end{enumerate}

\smallskip
A \textit{second set of questions}  
is whether developing non-myopic approaches brings performance gains compared to state of the art approaches that are myopic?

The second part of experiments (Section \ref{expe_moris}), therefore, compares the best \textsc{Economy} approach with
\citep{mori2017early} which has state of the art performance, as confirmed by a recent paper \citep{russwurm2019end}. 

Our experiments are designed to answer these two sets of questions.
This is why in Section \ref{datasets}, \textit{two different collections of datasets} are used in the experiments dedicated to each set of questions. 

\vspace{-5mm}
\subsection{Evaluation criterion}
\label{eval}
\vspace{-2mm}

In order to compare the methods, it is important to consider a criterion which expresses its worth for the final user.  
We define a new evaluation criterion used in our experiments both to optimize $K$ on a validation set and to evaluate the early classification approaches on a test set.
In actual use, ultimately, the value of employing an early classification method corresponds to the average cost that is incurred using it. For a given dataset ${\cal S}$, it is defined as follows:
{
\begin{equation}
AvgCost({\cal{S}}) \;  = \; \frac{1}{|\cal{S}|} \sum_{({\mathbf x_T}, y) \in {\cal{S}}}  \left ( \mathrm{C}_m \left ( h_{\widehat{t^*}}({\mathbf x_{\widehat{t^*}}})|y \right )  \; + \; \mathrm{C}_d(\widehat{t^*})    \right )
\label{eq:eval-cost}
\end{equation}}
where $\widehat{t^*}$ is the decision time chosen by the method as the one optimizing the trade-off between earliness and accuracy.
In our experiments, $AvgCost$  is evaluated for each dataset and for each early classification method. 
Statistical tests allow us to detect significant difference in performance. 

\subsection{Datasets}
\label{datasets}

Two distinct collections of datasets are used in our experiments.  

\vspace{-2mm}

\paragraph{The datasets for the comparison of the \textsc{Economy} approaches:}
~\\
With respect to the first set of questions presented in Section \ref{goal_expe}, about the role of the various design choices, one cannot easily measure differences of performance if the datasets only include time series that are easy to classify very early or that are hard to classify even when the whole series are known. Indeed, if this happens, all methods either decide early to output a label or wait until the end, and their performances are almost indistinguishable. In order, then, to be able to measure differences between the various online decision methods, we excluded datasets with these characteristics.

All the selected datasets come from the UEA \& UCR Time Series Classification Repository\footnote{ Available at : \url{http://www.timeseriesclassification.com}}. 
First, we removed potentially correlated datasets since it is important to select only independent datasets for the use of statistical tests.
We identified almost identical dataset names and sizes. 
For instance, the datasets ``Ford A'' and ``Ford B''  contain the same number of time series with the same length. 
In this case, we keep only one dataset chosen at random. 
Then, we learned a collection of classifiers for the remaining datasets and manually removed those for which the successive classifiers did not improve their quality over time. More specifically, we plotted the Cohen's \textit{kappa} score \citep{cohen1960coefficient} for each possible lengths of the input time series. We removed the datasets with low and almost constant quality of classifiers over time. 
Please note that the removed datasets are overwhelmingly the smallest, as the low number of training examples generally leads to poor quality classifiers. 
The 34 selected datasets and their description are available at \url{https://tinyurl.com/ycmbxurq}.

\paragraph{The datasets for comparisons with the state of the art methods:}
~\\ 
In order to be able to make direct comparison with the method described in \citep{mori2017early} we use the same datasets as they did. 
This benchmark consists of 45 datasets of variable sizes that come from a variety of application areas.  
This collection of datasets has also been used in \citep{schafer2020teaser} and  \citep{mori2019early}, making our experiences  easily comparable to previous works. 

\paragraph{Dataset preparation:}
First, the training and test sets were merged for each dataset to overcome the possibly unbalanced or biased split of the original data files. 
The remaining datasets were then transformed into binary classes since \textsc{Economy-$\gamma$} and \textsc{Economy-$\gamma$-lite} are limited to binary classification.
This was done by retaining the majority class and merging all the others.
In order to reduce the computation time of the experiments and to compare datasets with time series of different lengths, we trained a classifier every 5\% of the total length of the time series, instead of one classifier per time step, as done in \citep{mori2017early}. 
Furthermore, for each dataset and for each possible length (i.e \textit{5\%, 10\%, 15\% ...} of the total length), we extracted 60 features\footnote{More details are available in: \url{https://docs.google.com/spreadsheets/d/13u7L_5IX3XxFuq_SnbOZF1dXQfcBB0wR3PXhvevhPYA/}} from the corresponding truncated time series in order to train the associated classifiers. 
To do this, we used the Time Series Feature Extraction Library \citep{tsfel}, which automatically extracts features on the statistical, temporal and spectral domains.

\vspace{-4mm}
\subsection{Experimental protocol}
\label{expe_protocol}
\label{experimental_protocol}
\vspace{-2mm}
 
The datasets were divided by uniformly selecting 70\% of the examples for the training set and using the remaining 30\% for the test set. 
Furthermore,  
the training sets were divided into three disjoint subsets as follows: (subset a) 40\% for training  the various classifiers $\{h_t\}_{t \in \{1, \ldots, T\}}$; (subset b) 40\% for learning the meta parameters; (subset c) 20\% to optimize the number of groups in $\G$. 

\smallskip
\noindent
(subset a) \textit{Learning the collection of classifiers}: for each dataset, the classifiers corresponding to the possible lengths of the input time series (i.e. every 5\% of the total length) were learned.
The XGboost Python library{ \footnote{XGBoost is available in: \url{https://xgboost.readthedocs.io}}} was used, keeping the default values for the hyper-parameters.  

\smallskip
\noindent
(subset b) \textit{Learning the meta-parameters}: they were learned for each  \textsc{Economy} approach, except the parameter $K$ which is optimized using (subset c). For instance, a meta-model learned by the \textsc{Economy-$\gamma$} approach consists of: (i) the discretization into $K$ intervals of the confidence level for each classifier (one for each possible length), and (ii) the transition matrices between a time step to the next one (i.e. every 5\% of the time series length).  
 
\smallskip
\noindent
(subset c) \textit{Optimizing the number of groups $K$}:
the \textsc{Economy} algorithms were trained by varying the number of groups between $1$ to $20$ and evaluated by the $AvgCost(.)$ criterion which represents the average cost actually paid by the user (see Equation (\ref{eq:eval-cost})). The value of $K$ which minimizes the $AvgCost(.)$ criterion has been kept.
	
\vspace{-4mm}
\paragraph{Costs setting:}   
the misclassification cost was set in the same way for all datasets:  
$\mathrm{C}_m (\hat{y} |y)= 1$ if $\hat{y} \neq y$, and $= 0$ otherwise. The delay cost $\mathrm{C}_d(t)$ is provided by the domain experts in actual use case. In the absence of this knowledge, we define it as a linear function of time, with coefficient, or slope, $\alpha$:

{
\begin{equation}
\mathrm{C}_d(t) = \alpha \times \frac{t}{T} 
\label{eq:time-cost}
\end{equation}} 

The larger the $\alpha$ coefficient, the more costly it is to wait for more measurements in the incoming time series. 
The delay cost $\mathrm{C}_d(t)$ is obviously of paramount importance to control the best decision time. If $\alpha$ is very low, it does not hurt to wait for the whole time series to be known and $t^* = T$. If, on the contrary, $\alpha$ is very high, the gain in misclassification cost obtained thanks to more observations cannot compensate for the increase of the delay cost, and it is better to make a decision at the beginning of the observations.
Our experiments were run over a three ranges of values of $\alpha$: low time cost with $\alpha \in$ [1e-04, 2e-04, 4e-04, 8e-04, 1e-03, 3e-03, 5e-03, 8e-03]; medium time cost with $\alpha \in$  [0.01, 0.02, 0.03, 0.04, 0.05, 0.06, 0.07, 0.08, 0.09]; high time cost with $\alpha \in$ [0.1, 0.2, 0.3, 0.4, 0.5, 0.6, 0.7, 0.8, 0.9, 1.0]. 

\section{{Best design choices for \textsc{Economy}}}
\label{sec_results}

\vspace{-2mm}
This section presents the results of the experiments aimed at identifying the best design choices for the \textsc{Economy} approaches (see Section \ref{datasets}). 

\vspace{-4mm}
\subsection{Comparison of the \textsc{Economy} approaches with a non adaptive baseline }
\vspace{-2mm}

As a first sanity test, it is interesting to see if the \textsc{Economy} algorithms do indeed adapt the decision time to the incoming time series, or if they treat them all the same. In order to perform this test, each of the \textsc{Economy} approaches is run once in its adaptive mode, and once made unable to adapt by forcing the number of groups $K = 1$ (there is thus no difference made between the series). 

\begin{figure}[htbp!]
\centering
 \includegraphics[width=0.55\linewidth]{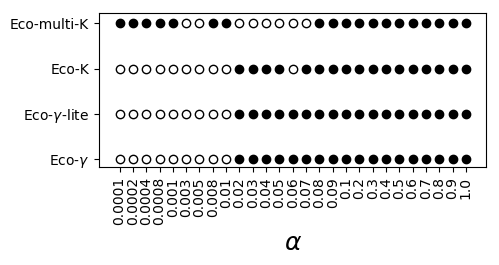}
\caption{Success of adapting the trigger times - Wilcoxon signed-rank test results for different values of $\alpha$: black dotes indicate success and circles failures.}
\label{fig:rep_adaptation_check}
\end{figure}

The \textsc{Economy} approaches are trained on the $34$ selected datasets by varying the value of $\alpha$, and then, evaluated on the test sets using the $AvgCost$ criterion.  
The Wilcoxon signed-rank test is used to assess whether the observed performance gap is significant. 
Figure \ref{fig:rep_adaptation_check} presents the results of the Wilcoxon signed-rank test for each \textsc{Economy} approach, applied over the $34$ datasets by varying the values of $\alpha$. 

In the range $\alpha \in [0.0001, 0.01]$, \textsc{Economy-multi-K} is the only approach that succeed in adapting its trigger times. By contrast, in the range $\alpha \in [0.02, 0.1]$, it appears that the \textsc{Economy} approaches actually succeed in adapting their trigger times, with the exception of \textsc{Economy-multi-K} which fails this test one-third of the time and behaves rather erratically when $\alpha$ varies. 

At the end, these approaches succeed in improving performance by adapting their trigger times, differing in their range of success.

\vspace{-3mm}
\subsection{Comparison of the \textsc{Economy} approaches}
\label{expe_ECO_only}

\paragraph{(a) Comparison with respect to the average decision cost}~\\
The $AvgCost$ criterion was evaluated on the $34$ test sets, and $\alpha$ was adjusted for each dataset in order to reveal the greatest differences in performance between the best and worst approach (see Table \ref{table:raw_result} for more details). 
The Nemenyi test \citep{Nemenyi62} was used to rank the different \textsc{Economy} approaches in terms of average decision cost.
The Nemenyi test consists of two successive steps.  
First, the Friedman test is applied to the average decision cost of competing approaches to determine whether their overall performance is similar. 
If not, the post-hoc test is applied to determine groups of approaches whose overall performance is significantly different from that of the other groups. 

\begin{figure}[htbp!]
\centering
\subfloat[]{\raisebox{3mm}{\label{fig:eco_comp_score:nemeny} \includegraphics[width=0.65\linewidth]{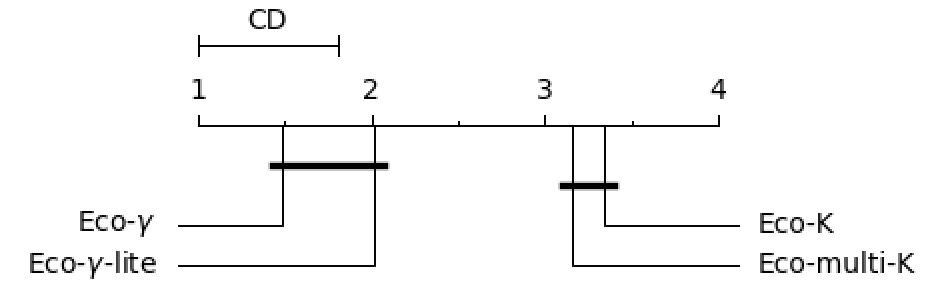}}}
\subfloat[]{\label{fig:eco_comp_score:wilcoxon} \includegraphics[width=0.25\linewidth]{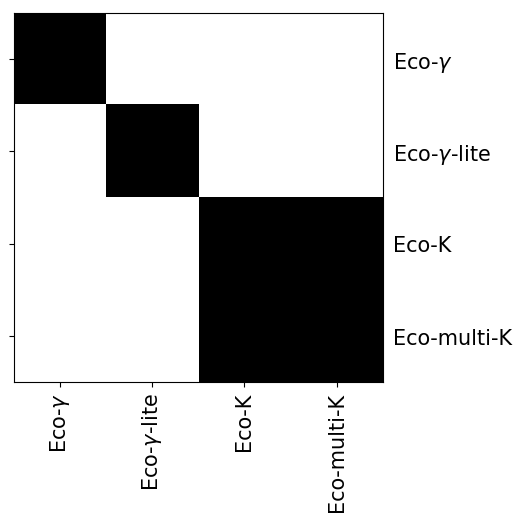}} 
\caption{Evaluation based on $AvgCost$: (a) Nemenyi test applied to the $34$ datasets; (b) pairwise comparison using the Wilcoxon signed-rank test, with black squares identifying non-significant comparisons.}
\label{fig:eco_comp_score}
\end{figure}

Figure \ref{fig:eco_comp_score:nemeny}, reporting the results of the Nemenyi test, shows two groups of methods of which the performances are significantly different. 
Specifically, the \textsc{Economy-$\gamma$} and \textsc{Economy-$\gamma$-lite} methods exhibit much better average decision costs than \textsc{Economy-K} and \textsc{Economy-multi-K}.

Figure \ref{fig:eco_comp_score:wilcoxon} shows pairwise comparison using the Wilcoxon signed-rank test  between the approaches. The small black squares identify pairs of approaches that do not differ significantly in performance. 
It is thus apparent that  \textsc{Economy-$\gamma$} performs significantly better than \textsc{Economy-$\gamma$-lite}.  

\begin{figure}[htbp!]
\centering
\subfloat[]{\raisebox{3mm}{\label{fig:eco_comp_prec:nemeny} \includegraphics[width=0.65\linewidth]{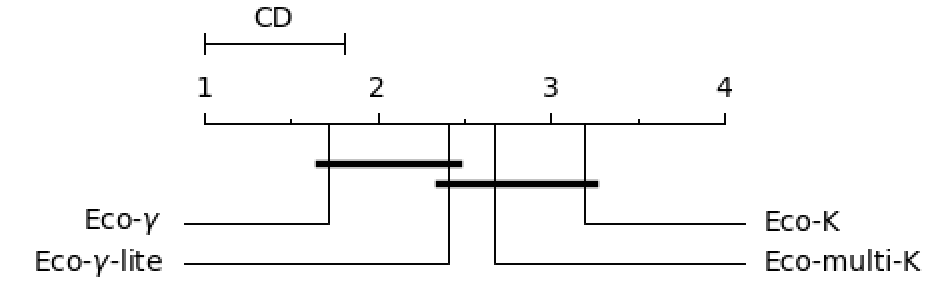}}}
\subfloat[]{\label{fig:eco_comp_prec:wilcoxon} \includegraphics[width=0.25\linewidth]{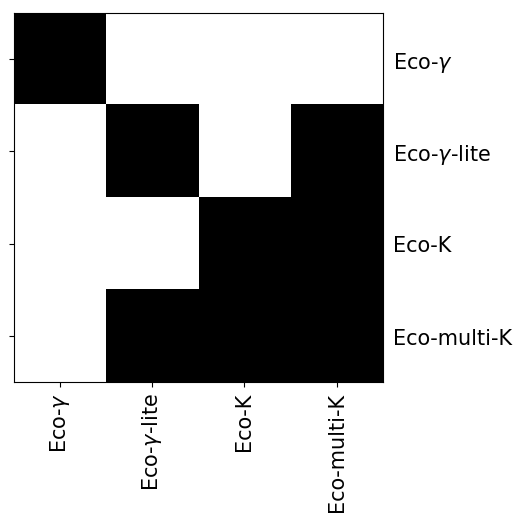}} \\
\subfloat[]{\raisebox{3mm}{\label{fig:eco_comp_kappa:nemeny} \includegraphics[width=0.65\linewidth]{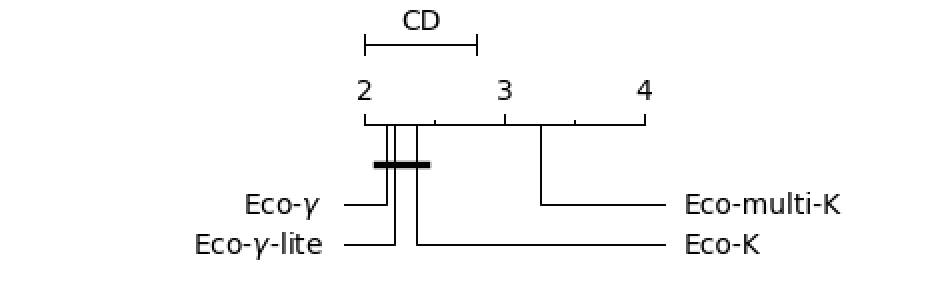}}}
\subfloat[]{\label{fig:eco_comp_kappa:wilcoxon} \includegraphics[width=0.25\linewidth]{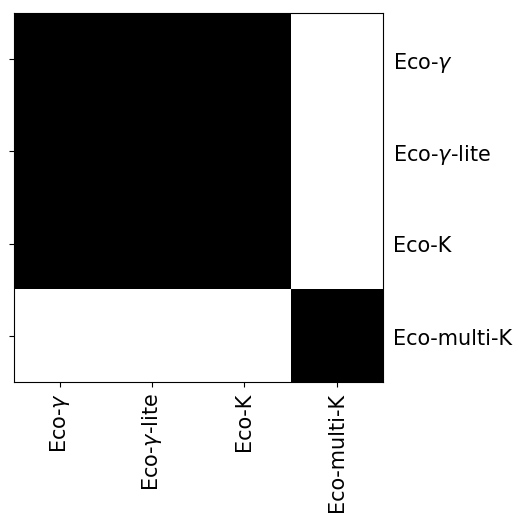}}
\caption{Earliness (a, b) and predictive performance (c, d) comparison of the \textsc{Economy} approaches.}
\label{fig:eco_comp}
\end{figure}

\paragraph{(b) Comparison with respect to the earliness of the decision time}~\\
In the following, the \textit{earliness} of early classification approaches is evaluated using the median of the trigger times normalized by the length of the series, defined by $ earliness = {med\{\widehat{t^*}\}} / {T} $ (see Table \ref{table:raw_result}). 
Figure \ref{fig:eco_comp_prec:nemeny} shows that \textsc{Economy-$\gamma$}, on average, triggers its decision earlier than  the competing methods, followed by \textsc{Economy-$\gamma$-lite}.
Furthermore, according to the Wilcoxon signed-rank test, this difference is significant compared to the other \textsc{Economy} approaches (Figure \ref{fig:eco_comp_prec:wilcoxon}).  
\paragraph{(c) Comparison with respect to the predictive performance of the algorithms}~\\
The \textit{predictive} performance is evaluated using the Cohen's \textit{kappa} score \citep{cohen1960coefficient} computed at $\widehat{t^*}$, since this criterion properly manages unbalanced datasets (see Table \ref{table:raw_result}). 
Again, the \textsc{Economy-$\gamma$} and \textsc{Economy-$\gamma$-lite} dominate in terms of predictive performance, but here the difference is not statistically significant. 

\begin{figure}[htbp!]
\centering
 \includegraphics[width=0.68\linewidth]{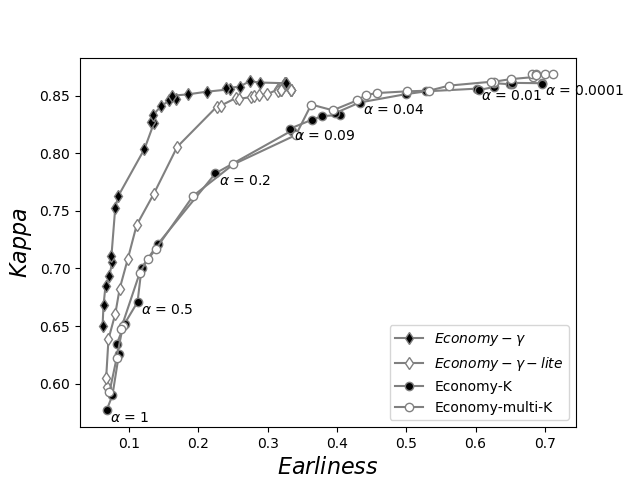}
\caption{Average \textit{Earliness} vs. Average \textit{Kappa} score obtain over the $34$ datasets by varying the slope of the time cost, such as $\alpha \in [10^{-4}, 1]$. }
\label{fig:eco_comp_pareto_all}
\end{figure}

\paragraph{(d) Pareto curves when varying the $\alpha$ coefficient controlling the delay cost}~\\
In Figure \ref{fig:eco_comp_pareto_all}, the coordinates of each point are given by the average \textit{Kappa} score and the average \textit{earliness} obtained over the $34$ datasets when the delay cost $\alpha$ is chosen in the range $[10^{-4}, 1]$, and the Pareto curve is drawn for each of the approaches. The result is strikingly clear. For each value of $\alpha$, \textsc{Economy-$\gamma$} dominates all others approaches, even if \textsc{Economy-$\gamma$-lite} is not far behind. The \textsc{Economy-K} and \textsc{Economy-multi-K} approaches yield much weaker results and are indistinguishable from each other. 

\begin{figure}[htbp!]
\centering
\subfloat[]{\raisebox{3mm}{\label{fig:eco_comp_diffscore:nemeny} \includegraphics[width=0.65\linewidth]{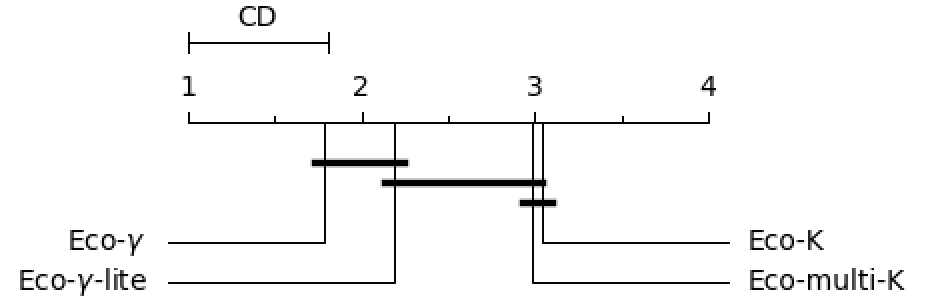}}}
\subfloat[]{\label{fig:eco_comp_diffscore:wilcoxon} \includegraphics[width=0.25\linewidth]{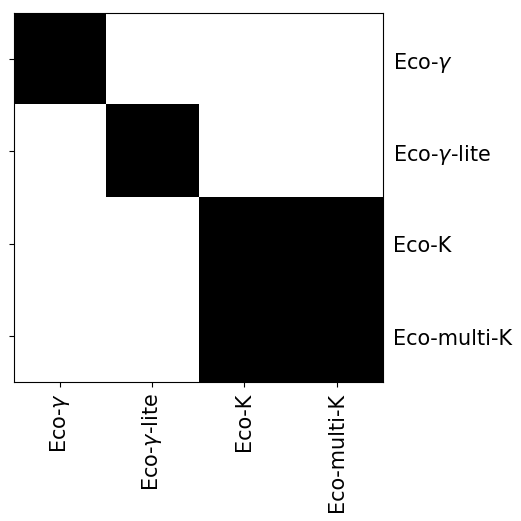}} 
\caption{Evaluation of the quality of online decisions based on $\Delta_{cost}$.}
\label{fig:eco_comp_diffscore}
\end{figure}

\paragraph{(e) Comparison with the best possible performance }~\\
The approaches presented are able to adapt their decision time $\widehat{t^*}$ to the characteristics of the time series and to perform well in terms of average decision costs $AvgCost$, but to what extent these results differ from the optimal ones $AvgCost^*$ computable \textit{after} the entire time series is known? 
For each dataset,  $\Delta_{cost} =  | AvgCost - AvgCost^* | $ was computed.
Figure \ref{fig:eco_comp_diffscore} shows that \textsc{Economy-$\gamma$} provides the best online decisions compared to the optimal ones, on average, followed by \textsc{Economy-$\gamma$-lite}.
According to the Wilcoxon signed-rank test, this difference is significant compared to the other \textsc{Economy} approaches (Figure \ref{fig:eco_comp_diffscore:wilcoxon}).

From all these results, several conclusions can be drawn. 

\begin{enumerate}
\setlength\itemsep{0.9em}

   \item The \textsc{Economy} approaches that partition the time series using the learned classifiers (supervised-based methods) perform significantly better than those which exploit the K-means algorithm (unsupervised-based methods).

   \item For the unsupervised approaches, partitioning the time series on full-length time series  (\textsc{Economy-K}), or on truncated ones (\textsc{Economy-multi-K}) does not significantly affect the performances. 

   \item Regarding the supervised methods, using a more sophisticated anticipation mechanism of the incoming time series as done by \textsc{Economy-$\gamma$} is profitable and allows it to beat the less sophisticated \textsc{Economy-$\gamma$-lite} method

\end{enumerate}

\vspace{-2mm}
\section{{Comparing \textsc{Economy-$\gamma$} and the state of the art}}
\label{sec_results_moris}
\vspace{-2mm}

This section compares \textsc{Economy-$\gamma$} to the state of the art approach \citep{mori2017early} and investigates the effects of the non-myopic property on its performance.   

\subsection{Comparing the performances}
\label{expe_moris}

An important question is whether it is worth considering explicitly, in a single optimization criterion, earliness and accuracy, as in the \textsc{Economy} approaches, and furthermore to adopt a non-myopic strategy with the modeling and computational costs involved. To assess this, we compared the \textsc{Economy} methods with a competing algorithm, called \textsc{SR} presented in \citep{mori2017early} which is claimed to dominate all other algorithms over 45 benchmark datasets.
The \textsc{SR} algorithm uses a trigger function to decide if the current prediction is reliable (output 1) or if it is preferable to wait for more data (output 0). Among several triggered functions, all of a heuristic nature, the most effective is:

{
\begin{equation}
Trigger \left ( h_t({\mathbf x}_t) \right )  =  \left\{
    \begin{array}{ll}
        0 & \mbox{if } \gamma_1 p_1 + \gamma_2 p_2 + \gamma_3 \frac{t}{T} \le 0 \\
        1 & \mbox{otherwise}
    \end{array}
\right.
\label{eq:mori_trigger_function}
\end{equation}} 
where $p_1$ is the largest posterior probability estimated by the classifier $h_t$: $p_1 = \operatornamewithlimits{ArgMax}_{y \in \cal Y} (\hat{p}(y|{\mathbf x}_t))$, $p_2$ is the difference between the two largest posterior probabilities, defined as $| \hat{p}(y=1|{\mathbf x}_t) - \hat{p}(y=0|{\mathbf x}_t) |$ in the case of binary classification problems, and where the last term $\frac{t}{T}$ represents the proportion of the incoming time series that is visible at time $t$.  

The parameters $\gamma_1, \gamma_2, \gamma_3$ are real values in $[-1, 1]$ to be optimized. In our experiments, these parameters were tuned using a grid-search over the set of values [$ -1, -0.95, -0.90 , ..., 0, 0.05, ..., 0.90, 0.95, 1 $] in order to minimize the criterion $AvgCost$.  
The optimization was carried out for all possible time cost functions with a slope  $\alpha \in [10^{-4}, 1]$.

\begin{figure}[htbp!]
\centering
\subfloat[]{\raisebox{3mm}{\label{fig:eco_comp_score_mori:nemeny} 
\includegraphics[width=0.5\linewidth]{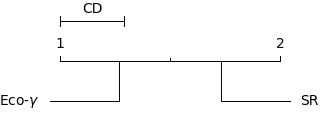}}}
\subfloat[]{\label{fig:eco_comp_score_mori:wilcoxon}  
\includegraphics[width=0.25\linewidth]{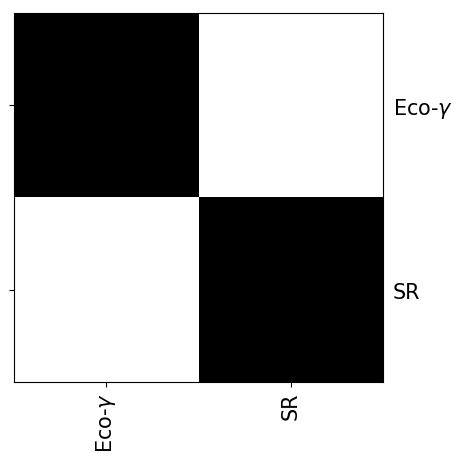}} 
\caption{\textsc{SR} \textit{vs.} \textsc{Economy-$\gamma$}: evaluation based on $AvgCost$.}
\end{figure}

After training, the $AvgCost$ criterion was evaluated on the 45 test sets, and $\alpha$ was adjusted for each dataset in order to find the most favorable setting for the $\textsc{SR}$ algorithm, namely one maximizing $AvgCost_{{SR}} - AvgCost_{{Eco-\gamma}}$. 

Figure \ref{fig:eco_comp_score_mori:nemeny} reports the results of the Nemenyi test and demonstrates that even in these situations favoring the SR algorithm, \textsc{Economy-$\gamma$} reaches significantly better performances. 
The Wilcoxon signed-rank test presented in Figure \ref{fig:eco_comp_score_mori:wilcoxon} reinforces this conclusion. 

\begin{figure}[htbp!]
\centering
\includegraphics[width=0.95\linewidth]{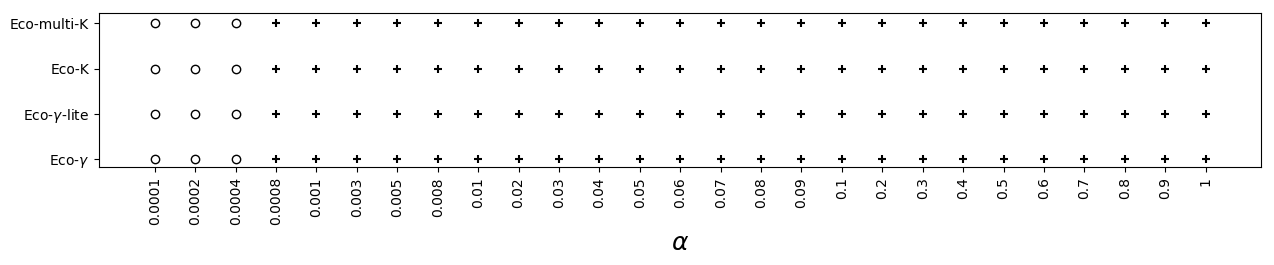}
\caption{$\textsc{Mori}$ \textit{vs.} \textsc{Economy} approaches : evaluation based on $AvgCost$ using the Wilcoxon signed-rank test, for different values of $\alpha$: ``$+$'' indicate success of \textsc{Economy} approaches and ``$\circ$'' insignificant difference in performance.}
\label{fig:test_wilcoxon_moris_vs_economy}
\end{figure}

We also carried out the Wilcoxon signed-rank test to compare the $\textsc{SR}$ approach with the four \textsc{Economy} approaches, for each value of $\alpha \in [10^{-4},1]$. The results (see Figure \ref{fig:test_wilcoxon_moris_vs_economy}) shows forcibly that the  \textsc{Economy} approaches perform significantly better than the $\textsc{SR}$ approach, regardless of the value of $\alpha$; except for $\alpha \in \{10^{-4}, 2.10^{-4}, 4.10^{-4}\}$ where this difference is not significant. 

\subsection{Measuring the effect of the non-myopic property of \textsc{Economy}}
\label{expe_moris}

One important feature that differentiates the \textsc{Economy} approaches from the state of the art methods is being non-myopic. Where standard methods decide whether to make a prediction at the current time based only on currently available information, the Economy algorithms look at future instants in order to predict the best decision time. 

This section presents experiments aimed at answering the following question: Is it better to be non-myopic for an online decision system?

\begin{figure}[htbp!]
 	\centering
 	\subfloat[]{\raisebox{3mm}{\label{fig:eco_comp_horizon_one:nemeny} \includegraphics[width=0.65\linewidth]{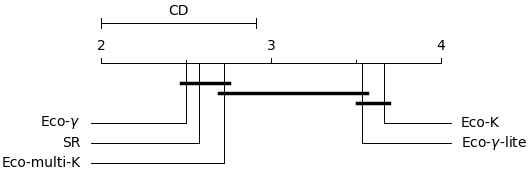}}}
 	\subfloat[]{\label{fig:eco_comp_horizon_one:wilcoxon} \includegraphics[width=0.25\linewidth]{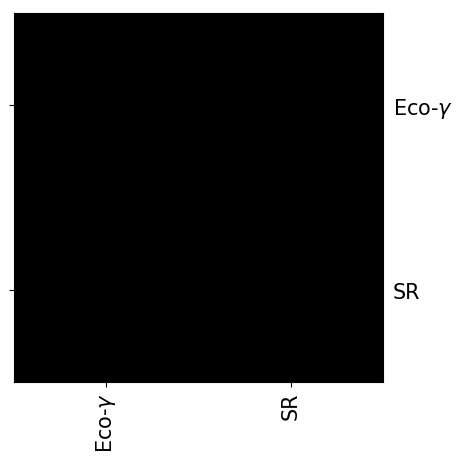}} 
 	\caption{Evaluation of the myopic version of \textit{Economy} approaches and SR approach based on $AvgCost$: (a) Nemenyi test applied to the $45$ datasets; (b) pairwise comparison using the Wilcoxon signed-rank test, with black squares identifying non-significant comparisons.}
 	\label{fig:eco_comp_horizon_one}
 \end{figure}

To answer this question, four myopic versions of the proposed \textsc{Economy} approaches were implemented by limiting the  horizon to only one measurement in the future, instead of looking at all the future time steps. The experiments performed with these myopic approaches are similar to those described in Section \ref{sec_results_moris}. The results are reported in Figure \ref{fig:eco_comp_horizon_one}.

Figure \ref{fig:eco_comp_horizon_one:nemeny} reports the results of the Nemenyi test between the myopic \textsc{Economy} approaches and the SR approach of \citep{mori2017early}.
They are cast in the same group. Moreover, the Wilcoxon signed-rank test presented in Figure \ref{fig:eco_comp_horizon_one:wilcoxon} shows that there is no statistical significant difference in performance. Consequently, it appears that the non-myopic feature is a key property required to obtain better results.

\section{Conclusions}
\label{sec_conclusion}

An increasing number of applications require the ability to recognize the class of an incoming time series as quickly as possible  without unduly compromising the accuracy of the prediction. In this paper, we reformulated in a generic way 
an optimization criterion put forward in \citep{dachraoui2015early} which takes into account both the cost of misclassification and the cost of delaying the decision. 

This generic framework has been technically declined, leading to the design of three new ``non-myopic'' algorithms - i.e. able to anticipate the expected future gain in information in balance with the cost of waiting. 
In one class of algorithms, unsupervised-based, the expectations use the clustering of time series, while in a second class, supervised-based, time series are grouped according to the confidence level of the classifier used to label them. 

We have defined a new evaluation criterion that represents the average cost incurred when the method is applied over a set of labelled time series. This criterion makes it possible to evaluate both earliness and predictive performance as a single objective, with respect to the ground truth. It offers a well-grounded framework widely applicable for the comparison of methods. 

Extensive experiments carried out on real datasets using a large range of delay cost functions show that the presented algorithms are able to satisfactorily solve the earliness vs. accuracy trade-off, with the supervised {partition} based approaches generally faring better than the unsupervised {partition} based ones. In addition, all these methods perform better in a wide variety of conditions than the state of the art competitive method of \cite{mori2017early}. 
{The non-myopic feature of the \textsc{Economy} approaches is required for this good achievement.}
{We have shown that the non-myopic property of the \textsc{Economy} approaches plays a key role for these good performances.}

Given the merit of the proposed approach, we envision several extensions. One is to allow the {\textsc{Economy-$\gamma$-lite} and \textsc{Economy-$\gamma$}} approaches, which use the confidence level of a binary classifier, to solve multi-classes problems. A second one is to use a supervised clustering technique to compute groups of times series (see \cite{lemaire2020predictive}) {in the \textsc{Economy-K} and \textsc{Economy-multi-K} approaches}. Finally, we are working on the adaptation of  these methods to the on-line detection of anomalies in a data stream.

\section*{Acknowledgements}
We thank Vincent Lemaire and Fabrice Clérot (Orange Labs) for their advice and interesting discussions. We also thank Orange Labs for supporting this research.

\bibliographystyle{spbasic}

\bibliography{biblio}

\begin{table*}\centering
	\scriptsize   
	\begin{turn}{90}
		\setlength{\tabcolsep}{1.0pt}
		\begin{tabular}{@{}lcllllclllllcllllcllll@{}}
			\toprule
			
			\multirow{2}{*}{Datasets} & \multirow{2}{*}{\text{$\alpha$}}
			& \multicolumn{4}{c}{Nb groups: K } \phantom{abc} &
			& \multicolumn{4}{c}{$AvgCost$} & \phantom{abc}&
			
			\multicolumn{4}{c}{$Median$ $\widehat{t^*}$} & \phantom{abc} &
			\multicolumn{4}{c}{$Kappa$}
			\\\cmidrule{3-6} \cmidrule{8-11} \cmidrule{13-16} \cmidrule{18-21}&
			
			{\text{~~~~~~~~~}} & E-$\gamma$ & E-$\gamma$-l & E-K & E-m-K && \text{Eco-$\gamma$~~} & \text{Eco-$\gamma$-l} & \text{Eco-K~~} & \text{Eco-m-K} && \text{Eco-$\gamma$~~} & \text{Eco-$\gamma$-l} & \text{Eco-K} & \text{Eco-m-K} && \text{Eco-$\gamma$~~} & \text{Eco-$\gamma$-l} & \text{Eco-K} & \text{Eco-m-K} \\
			\toprule
			
			$CBF$ & 0.8 & 11 & 16 & 7 & 4 &&  0.167 & \textbf{0.163} & 0.231 & 0.222 &&             \textbf{0.09}& 0.14& 0.23& 0.23&& 0.83& \textbf{0.89}& 0.87& \textbf{0.89} \\
			$ChlorineConcentration$ & 0.4 & 14 & 11 & 20 &  1 && \textbf{0.252}& 0.302& 0.303& 0.322&&       \textbf{0.10}& 0.14& 0.05& 0.14&& \textbf{0.59}& 0.50& 0.46& 0.46\\
			$CinCECGTorso$ & 0.1 & 20 & 15 & 1 & 1 && 0.014& \textbf{0.012}& 0.021& 0.021&&       0.05& 0.05& 0.05& 0.05&& \textbf{0.98}& \textbf{0.98}& 0.96& 0.96\\
			$Crop$ & 0.06 &10 & 19 & 19 & 18 && \textbf{0.028}& 0.031& 0.031& 0.042&&             0.04& 0.04& 0.04& 0.04&& 0.70& 0.68& 0.65& \textbf{0.71}\\
			$ECG5000$ & 0.5 & 18 & 15 & 1 & 19 &&   \textbf{0.046}& 0.051& 0.065& 0.064&&           0.05& 0.05& 0.05& 0.05&& \textbf{0.97}& 0.96& 0.92& 0.92\\
			$ECGFiveDays$ &  0.3 & 5 & 7 & 14 & 7 &&    0.115&\textbf{0.097}& 0.163& 0.125&&     \textbf{0.09}& 0.13& 0.49& 0.18&& 0.83& 0.92& \textbf{0.95}& 0.92\\
			$ElectricDevices$ & 0.1 & 18 & 4 & 13 & 10 &&  0.114& \textbf{0.109}& 0.150& 0.146&&          0.08& \textbf{0.04}& 0.83& 0.46&& 0.75& 0.80& \textbf{0.82}& 0.75\\
			$FaceAll$ & 0.01 & 16 & 16& 19 & 16 &&  \textbf{0.002}& \textbf{0.002}& 0.006& 0.004&&              \textbf{0.05}&\textbf{0.05}&\textbf{0.05}& 0.32&& \textbf{0.99}& \textbf{0.99}& 0.98& \textbf{0.99}\\
			$FacesUCR$ & 0.5 & 11 & 2 & 13 & 18 &&  \textbf{0.093}& 0.111& 0.123& 0.133&&          \textbf{0.05}& \textbf{0.05}& 0.14& 0.14&& \textbf{0.77}& 0.67& 0.69& 0.72\\
			$FiftyWords$ &  0.5 & 15 & 2 & 13 & 7 && \textbf{0.104}& 0.119& 0.148& 0.160&&           \textbf{0.05}& 0.10& \textbf{0.05}& \textbf{0.05}&& \textbf{0.68}& 0.66& 0.29& 0.55\\
			$FordA$ & 0.3 & 14 & 7 & 7& 2 &&  \textbf{0.114}& 0.118& 0.155& 0.164&&                0.15& 0.15& 0.15& 0.15&& \textbf{0.88}& \textbf{0.88}& 0.81& 0.79\\
			$FreezerRegularTrain$  &  0.2 & 17 & 4 & 7& 17 && \textbf{0.016}& 0.021& 0.024& 0.025&&          0.05& 0.05& 0.05& 0.05&& \textbf{0.99}& 0.98& 0.98& 0.98\\
			$HandOutlines$ & 0.01 & 17 & 4 & 4 & 10 &&  \textbf{0.109}& 0.113& 0.124& 0.160&&            \textbf{0.25}& 0.40& 0.50& 1.00&& \textbf{0.76}& \textbf{0.76}& 0.73& 0.66\\
			$InsectWingbeatSound$ & 1 & 19 & 1 & 1 & 1 && 0.151& \textbf{0.130}& \textbf{0.130}& \textbf{0.130}&&            0.05& 0.05& 0.05& 0.05&& 0.25& \textbf{0.29}& \textbf{0.29}& \textbf{0.29}\\
			$ItalyPowerDemand$ & 0.5 & 9 & 14 & 1 & 1 &&  \textbf{0.265}& 0.302& 0.349& 0.349&&             \textbf{0.04}& 0.08& 0.33& 0.33&& 0.61& 0.54& \textbf{0.64}& \textbf{0.64}\\
			$Mallat$ & 0.08 & 9 & 12 & 10 & 17 &&  \textbf{0.016}& 0.019& 0.034& 0.028&&                 \textbf{0.05}& \textbf{0.05}& 0.30& 0.15&& 0.97& 0.96& 0.93& \textbf{0.99}\\
			$MedicalImages$ & 0.07 & 12 & 11 & 19 & 20 &&  \textbf{0.210}& 0.238& 0.240& 0.279&&          \textbf{0.08}& 0.16& 0.12& 0.69&& \textbf{0.61}& 0.57& 0.55& 0.54\\
			$MelbournePedestrian$ &0.8 & 12 & 3 & 1 & 11 &&  0.126& \textbf{0.111}& 0.134& 0.122&&         0.04& 0.04& 0.04& 0.04&& 0.71& 0.85& 0.70& \textbf{0.86}\\
			$MixedShapesRegularTrain$ & 0.1 & 20 & 5 & 3 & 17 &&  0.030&\textbf{0.020}& 0.047& 0.044&&      \textbf{0.05}& 0.10& 0.15& 0.10&& 0.94& \textbf{0.97}& 0.91& 0.92\\
			$MoteStrain$ & 0.4 & 7 & 18 & 11 & 18 &&  \textbf{0.128}& 0.142& 0.156& 0.178&&        \textbf{0.10}& 0.14& 0.14& 0.29&& \textbf{0.86}& 0.84& 0.80& 0.82\\
			$NonInvasiveFatalECGT2$ & 0.04 & 5 & 19 & 9 & 7 &&   \textbf{0.011}& \textbf{0.011}& 0.014& 0.012&&              0.05& 0.05& 0.05& 0.05&& 0.81& 0.80& 0.71& \textbf{0.83}\\
			$PhalangesOutlinesCorrect$ & 0.2 & 5 & 16 & 4 & 1 &&   \textbf{0.297}& 0.348& 0.320& 0.316&&          \textbf{0.15}& 0.25& \textbf{0.15}& \textbf{0.15}&& \textbf{0.4}& 0.29& 0.36& 0.35\\
			$ProximalPhalanxOutlineCo$ &0.5 &  6 & 4 & 7 & 17 &&  \textbf{0.267}& 0.300& 0.300& 0.306&&            \textbf{0.05}& 0.10& \textbf{0.05}& 0.20&& \textbf{0.43}& 0.39& 0.36& 0.55\\
			$SemgHandGenderCh2$ &  0.3 & 4 & 8 & 1 & 7 && 0.176& 0.176& \textbf{0.174}& 0.265&&           \textbf{0.15}& 0.20& 0.20& 0.40&& 0.74& \textbf{0.77}& 0.76& 0.64\\
			$SonyAIBORobotSurface2$ & 0.8 & 7 & 10 & 17 & 15 &&  \textbf{0.176}& 0.210& 0.228& 0.208&&           \textbf{0.09}& 0.18& 0.18& 0.18&& 0.84& \textbf{0.85}& 0.82& 0.84\\
			$StarLightCurves$ & 0.3 & 17 &  12 & 18 & 3 &&  \textbf{0.062}& 0.067& 0.089& 0.095&&            \textbf{0.05}& \textbf{0.05}& 0.15& 0.20&& \textbf{0.93}& 0.92& 0.89& 0.91\\
			$Strawberry$ & 0.6 & 2 & 9 & 17 & 8 &&   0.195& \textbf{0.193}& 0.251& 0.201&&            0.09& 0.14& \textbf{0.05}& 0.19&& 0.78& 0.77& 0.59& \textbf{0.83}\\
			$Symbols$ &  0.2 & 15 & 15 & 15 & 4 && 0.034&\textbf{0.031}& 0.057& 0.042&&            0.05& 0.05& 0.05& 0.05&& 0.97& 0.98& 0.89& \textbf{0.99}\\
			$TwoLeadECG$ & 0.9 & 15 & 16 & 19 & 19 &&  \textbf{0.160}& 0.168& 0.197& 0.185&&            \textbf{0.10}& \textbf{0.10}& 0.15& 0.15&& 0.88& \textbf{0.90}& 0.86& 0.86\\
			$TwoPatterns$ & 0.08 & 13 & 11 & 10 & 19 &&   0.079& \textbf{0.065}& 0.104& 0.104&&         \textbf{0.56}& 0.66& 0.89& 0.89&& 0.90& \textbf{0.95}& 0.91& 0.91\\
			$UWaveGestureLibraryX$ & 0.5 & 12 & 1 & 18 & 17 && \textbf{0.130}& 0.148& 0.153& 0.126&&      0.05& 0.05& 0.05& 0.05&& 0.42& 0.14& 0.22& \textbf{0.62}\\
			$Wafer$ & 0.2 & 18 & 18 & 15& 6 &&   \textbf{0.010}& \textbf{0.010}& 0.011& \textbf{0.010}&&         0.05& 0.05& 0.05& 0.05&& \textbf{1.00}& \textbf{1.00}& 0.99& \textbf{1.00}\\
			$WordSynonyms$ & 0.6 & 19 & 20 & 20 & 1 &&  \textbf{0.177}& 0.218& 0.209& 0.307&& \textbf{0.10}& \textbf{0.10}& \textbf{0.10}& 0.14&& \textbf{0.63}& 0.53& 0.54& 0.36\\
			$Yoga$ & 0.03 & 14 & 8 & 8 & 20 &&  \textbf{0.168}& 0.246& 0.183& 0.193&&           \textbf{0.10}& 0.49& 0.49& 0.49&& \textbf{0.67}& 0.54& 0.66& 0.65\\
	
			\bottomrule
			
		\end{tabular}
	\end{turn}
	
	\caption{Details of experimental results for each dataset.}
	\label{table:raw_result}
	
\end{table*}

\end{document}